\documentclass[11pt]{article}
\usepackage[margin=1in]{geometry}
\usepackage{amsmath,amssymb,amsthm}
\usepackage{cite}
\usepackage{hyperref}
\usepackage{float}
\usepackage{tikz}
\usetikzlibrary{positioning,arrows.meta,fit,backgrounds,shapes.geometric,matrix,calc}
\usepackage{xcolor}
\usepackage{graphicx}

\IfFileExists{latexml.sty}{\usepackage{latexml}}{\newif\iflatexml}

\iflatexml\renewcommand{\vspace}[1]{}\fi

\newcommand{\Mset}{\mathcal{M}}
\newcommand{\Oset}{\mathcal{O}}
\newcommand{\Xset}{\mathcal{X}}
\newcommand{\Yset}{\mathcal{Y}}
\newcommand{\Pset}{\mathcal{P}}
\newcommand{\Dset}{\mathcal{D}}
\newcommand{\Tset}{\mathcal{T}}
\newcommand{\Init}{\iota}
\newcommand{\R}{\mathbb{R}}

\newcommand{\pcoord}[1]{{\scriptsize\mdseries\textcolor{black!60}{#1}}}

\newcommand{\shapepanel}[6]{%
  \node[shapebox, #1] (#2) {};
  \node[font=\scriptsize\bfseries, anchor=north] at ([yshift=-1.2mm]#2.north) {#3};
  \node[slottag] at ([yshift=-6.5mm]#2.north) {#4};
  \node[expl, text width=31mm] at ([yshift=2.5mm]#2.south) {#5};
  \begin{scope}[shift={($(#2.center)+(0,-0.13)$)}] #6 \end{scope}%
}

\newcommand{\topopanel}[5]{%
  \node[topobox, #1] (#2) {};
  \node[font=\scriptsize\bfseries, anchor=north] at ([yshift=-1mm]#2.north) {#3};
  \begin{scope}[shift={($(#2.north)+(0,-14mm)$)}, scale=1.2] #5 \end{scope}%
  \node[topocap] at ([yshift=-23mm]#2.north) {#4};
}

\title{A Vocabulary for Multi-Agent Automated Research Systems}
\author{Bardiya Akhbari\\{\small Amazon AGI}\\{\small\textit{bardiyaa@amazon.com}}}
\date{}

\begin{document}
\maketitle

\vspace{-5mm}
\begin{abstract}
We introduce a vocabulary for automated research systems built from one or more agents to make their design choices easier to describe and compare. The vocabulary specifies who the agents are, which operations the harness exposes, who may invoke them, how agents communicate, what state is visible within and across runs, how the next action is chosen, how a run begins, and how outputs are scored. A trajectory records one run from the input task to the returned artifact. Because agents, operations, and initialization may be stochastic, repeated runs on the same task induce a distribution over trajectories rather than a single behavior.

The vocabulary assigns each structural design question, such as when agents should communicate, gain or lose a capability, or carry information across runs, to a distinct coordinate that can be varied on its own. It also makes the evaluator a component of the system, since reported gains depend on how closely the proxy score matches true quality. That separation also splits the vague complaint that these systems lack \emph{taste} into two failures with different fixes. Generative taste is the rate at which a system proposes novel trajectories before any score is observed, and evaluative taste is the gap between the proxy score and the quality it should match. We instantiate the vocabulary on recent autoresearch systems to illustrate that it covers designs that differ widely in structure.
\end{abstract}

\begin{figure}[H]
\centering
\iflatexml
\includegraphics[width=0.98\textwidth]{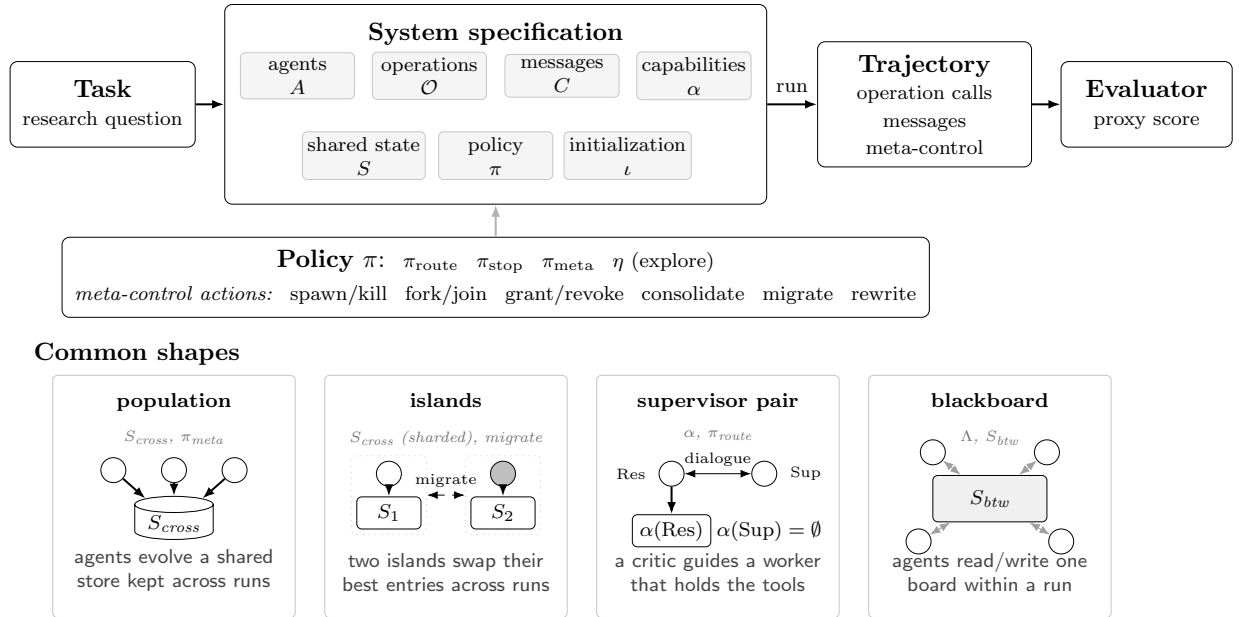}
\else
\resizebox{0.98\textwidth}{!}{%
\begin{tikzpicture}[
  font=\small,
  box/.style={draw, rounded corners=3pt, inner sep=5pt, align=center},
  stage/.style={box, minimum height=12mm},
  slot/.style={draw=gray!45, fill=gray!8, rounded corners=2pt, minimum height=6mm, minimum width=16mm, inner sep=2pt, align=center, font=\scriptsize},
  shapebox/.style={draw=gray!45, rounded corners=2pt, minimum width=34mm, minimum height=34mm, inner sep=2pt},
  slottag/.style={font=\tiny\itshape, text=black!55, anchor=north, align=center},
  isle/.style={draw=gray!40, rounded corners=2pt, dotted, inner sep=2pt},
  ag/.style={circle, draw, minimum size=3.5mm, inner sep=0pt},
  store/.style={draw, rounded corners=2pt, minimum width=9mm, minimum height=4.5mm, font=\scriptsize\itshape, inner sep=1pt},
  db/.style={draw, cylinder, shape border rotate=90, aspect=0.28, minimum width=11mm, minimum height=6mm, font=\scriptsize\itshape, inner sep=1pt},
  arr/.style={-{Latex[length=1.7mm]}, thick},
  darr/.style={{Latex[length=1.5mm]}-{Latex[length=1.5mm]}},
  expl/.style={font=\sffamily\scriptsize, text=gray!55!black, align=center, anchor=south},
]
\node[stage, minimum width=18mm] (task) at (0,0) {\textbf{Task}\\[-1pt]{\scriptsize research question}};
\node[box, minimum width=76mm, minimum height=28mm, right=4mm of task] (system) {};
\node[stage, minimum width=30mm, right=7mm of system] (traj) {\textbf{Trajectory}\\[-1pt]{\scriptsize operation calls}\\[-1pt]{\scriptsize messages}\\[-1pt]{\scriptsize meta-control}};
\node[stage, minimum width=24mm, right=4mm of traj] (score) {\textbf{Evaluator}\\[-1pt]{\scriptsize proxy score}};

\node[font=\small\bfseries, anchor=north] at ([yshift=-1.2mm]system.north) {System specification};
\node[slot] at ($(system.center)+(-2.78,0.4)$) {agents\\$A$};
\node[slot] at ($(system.center)+(-0.93,0.4)$) {operations\\$\Oset$};
\node[slot] at ($(system.center)+(0.93,0.4)$) {messages\\$C$};
\node[slot] at ($(system.center)+(2.78,0.4)$) {capabilities\\$\alpha$};
\node[slot] at ($(system.center)+(-1.85,-0.72)$) {shared state\\$S$};
\node[slot] at ($(system.center)+(0.0,-0.72)$) {policy\\$\pi$};
\node[slot] at ($(system.center)+(1.85,-0.72)$) {initialization\\$\Init$};

\draw[arr] (task) -- (system);
\draw[arr] (system) -- node[above=0.3mm, font=\scriptsize] {run} (traj);
\draw[arr] (traj) -- (score);

\node[box, below=4.5mm of system, minimum width=82mm, align=center] (actions) {%
  \textbf{Policy} $\pi$:\;
  {\scriptsize $\pi_{\text{route}}$ \enspace $\pi_{\text{stop}}$ \enspace $\pi_{\text{meta}}$ \enspace $\eta$ (explore)}\\[1pt]
  {\scriptsize\itshape meta-control actions:}\;
  {\scriptsize spawn/kill \enspace fork/join \enspace grant/revoke \enspace consolidate \enspace migrate \enspace rewrite}
};
\draw[arr, gray!60] (actions.north) -- (system.south);

\node[font=\small\bfseries, anchor=west] at (-1.1,-3.5) {Common shapes};

\shapepanel{at={(1.0,-5.5)}}{pop}{population}{$S_{\text{cross}}$, $\pi_{\text{meta}}$}{agents evolve a shared\\store kept across runs}{
  \node[ag] (p1) at (-0.85,0.5) {};
  \node[ag] (p2) at (0,0.5) {};
  \node[ag] (p3) at (0.85,0.5) {};
  \node[db] (ps) at (0,-0.25) {$S_{\text{cross}}$};
  \draw[arr] (p1) -- (ps);
  \draw[arr] (p2) -- (ps);
  \draw[arr] (p3) -- (ps);
}

\shapepanel{right=4mm of pop}{islands}{islands}{$S_{\text{cross}}$ (sharded), migrate}{two islands swap their\\best entries across runs}{
  \node[ag] (ia1) at (-0.8,0.45) {};
  \node[store] (i1) at (-0.8,-0.1) {$S_1$};
  \node[ag, fill=black!25] (ia2) at (0.8,0.45) {};
  \node[store] (i2) at (0.8,-0.1) {$S_2$};
  \draw[arr] (ia1) -- (i1);
  \draw[arr] (ia2) -- (i2);
  \begin{scope}[on background layer]
    \node[isle, fit=(ia1)(i1)] (isleA) {};
    \node[isle, fit=(ia2)(i2)] (isleB) {};
  \end{scope}
  \draw[darr, dashed] (isleA.east) -- node[font=\tiny, above=-0.2mm] {migrate} (isleB.west);
}

\shapepanel{right=4mm of islands}{dialogue}{supervisor pair}{$\alpha$, $\pi_{\text{route}}$}{a critic guides a worker\\that holds the tools}{
  \node[ag] (dr) at (-0.65,0.45) {};
  \node[ag] (ds) at (0.65,0.45) {};
  \node[font=\tiny, left=0.5mm of dr] {$\mathrm{Res}$};
  \node[font=\tiny, right=0.5mm of ds] {$\mathrm{Sup}$};
  \draw[darr] (dr) -- node[font=\tiny, above=-0.2mm] {dialogue} (ds);
  \node[store, minimum width=11mm] (tr) at (-0.65,-0.35) {$\alpha(\mathrm{Res})$};
  \draw[arr] (dr) -- (tr);
  \node[font=\scriptsize] at (0.72,-0.35) {$\alpha(\mathrm{Sup})=\emptyset$};
}

\shapepanel{right=4mm of dialogue}{blackboard}{blackboard}{$\Lambda$, $S_{\text{btw}}$}{agents read/write one\\board within a run}{
  \node[store, fill=gray!12, minimum width=16mm, minimum height=6.5mm] (bb) at (0,0.1) {$S_{\text{btw}}$};
  \foreach \i/\x/\y in {1/-0.8/0.75, 2/0.8/0.75, 3/-1/-0.5, 4/1/-0.5} {
    \node[ag] (b\i) at (\x,\y) {};
    \draw[darr, gray!75] (b\i) -- (bb);
  }
}
\end{tikzpicture}%
}
\fi
\caption{A multi-agent system is a set of components. A research task enters on the left. The system produces a trajectory, the full record of which agent acted, what it did, and what came back at each step. The evaluator grades that trajectory and returns a score. The lower row shows common shapes, each a different setting of those same components.}
\label{fig:overview}
\end{figure}

\section{Introduction}

Recent multi-agent large language model (LLM) systems automate research end-to-end: they collect and read prior work, propose ideas, write code, run experiments, interpret results, and iterate. Several of these systems frame their automation as explicit search over research trajectories. AIRA$_{2}$ searches a population of candidate experiments via evolutionary selection \cite{aira2026}. Glia searches a space of inference-system designs via a supervisor–researcher agent pair optimizing simulated latency \cite{hamadanian2025glia}. The Automated Alignment Researcher (AAR) searches for alignment algorithms by hill-climbing on a scalar reward signal \cite{wen2026automatedw2s}. These systems share a common shape, where each defines an objective, a search space of research artifacts, and an iterative strategy for navigating it. Lacking a common language to describe them, we formalize the shared structure as a framework for automated research.

These systems differ along many axes simultaneously. One differs from another in communication topology, another in cross-run memory, another in initialization, another only in the evaluator. A claim that a "multi-agent system" or an "autoresearcher" is better is ambiguous until we know which axis changed and why. A vocabulary should therefore decompose the system into coordinates fine enough to isolate each choice. Some decomposition decisions are non-obvious. We separate the operation universe (what can be done) from the capability assignment (who may do it), so that least-privilege and asymmetric designs are native rather than described by side comment. We include the evaluator as a primary component rather than treating it as external infrastructure, because reported gains depend on how closely the proxy score correlates with the true quality -- and that gap is often large \cite{hamadanian2025glia, wen2026automatedw2s}. We include initialization explicitly because the same architecture seeded differently can search entirely different regions of trajectory space.

Therefore, we formalize a multi-agent automated research system as the tuple
$$\Mset = \langle A, \Oset, C, \alpha, S, \pi, \Init, e \rangle,$$
where $A$ is the set of agents, $\Oset$ the operation universe (tools, skills, hooks, and other callable actions), $C=(\Lambda,\sigma)$ the communication structure ($\Lambda$ is the communication space, and $\sigma$ is the message protocol), $\alpha : A \to 2^{\Oset}$ the capability assignment, $S$ the shared state, $\pi$ the control policy, $\Init$ the initialization function, and $e$ the evaluator. In plain terms, the tuple records who is in the system, what can be done, who may do it, how agents communicate, what shared state exists, what runs when, how the run starts, and how the result is graded. We have split the operation universe and capability assignment deliberately. In our notation, $\Oset$ contains the invocable operations available to the system, such as application programming interfaces (APIs), shell commands, scripts, simulators, retrieval calls, or reusable skills loaded by the harness; $\alpha$ is the permission structure over that universe. The tuple formalizes what recent practice calls the \emph{harness}, the system layer around a base model that orchestrates its tool use, context, and control~\cite{weng2026harness}.

With this vocabulary a reader can identify which coordinate a comparison changed, and a designer can develop new techniques by varying one coordinate at a time. An improvement credited to ``multi-agent system'' may instead come from communication, initialization, cross-run memory, evaluator integrity, or capability assignment, and the tuple separates them into distinct components (Figure~\ref{fig:overview}). Treating the evaluator as a coordinate also gives us a handle on \emph{taste}, a notion left informal, which we split into whether the system proposes good candidates (generative taste) and whether its evaluator scores them faithfully (evaluative taste).

Our contribution is a language that describes these systems in common terms and identifies which coordinate each design changes. In the following sections, we define the problem space (Section~\ref{sec:problem}), the system specification (Section~\ref{sec:mas}), and the trajectory notation (Section~\ref{sec:trajectory}). We then discuss optimization challenges (Section~\ref{sec:objective}), apply the vocabulary to current autoresearch systems (Section~\ref{sec:case-studies}), and close with open design questions (Section~\ref{sec:discussion}). For related work, see Appendix~\ref{sec:related}.

\section{Problem specification}\label{sec:problem}

We can only judge a system relative to a problem, which we write as the tuple
\begin{equation}
\Pset = \langle \Xset, \Yset, \phi, q, \omega, B, \Dset \rangle.
\end{equation}
A problem defines the tasks a system faces, the artifacts that count as solutions, and how their quality is measured. A task $x \in \Xset$ (the input space) might be a research question, a bug report, a clinical case, or a code-improvement target, and a solution $y \in \Yset$ (the output space) is the corresponding artifact, such as a paper, a patch, a diagnosis, or an optimized program.

The \emph{task feature map} $\phi : \Xset \to \R^{d}$, for some $d \geq 1$, maps each task to a vector of measurable properties. For instance, the task descriptions ``write a data-loading script'' and ``discover a sorting algorithm and prove it'' are both tasks $x \in \Xset$, but $\phi$ separates them by coordinates such as the number of subtasks (one vs. several) and the performance of a fixed reference agent on similar tasks. Grouping tasks by these coordinates lets us reason about whole families at once.

The \emph{true quality} $q : \Xset \times \Yset \to \R^{k}$ is the ideal evaluation function. It returns the actual quality of $y$ on $x$ along $k \geq 1$ objective dimensions (e.g., $k=3$ for correctness, novelty, and cost). The scalarization $\omega:\mathbb{R}^k\to\mathbb{R}$ combines the $k$ scores into a single number, so solutions can be ranked. In practice, $q$ is uncomputable, and the discrepancy between a computable evaluator and $q$ is the proxy-quality gap. Each task has a hard \emph{budget}, $B \in \mathbb{R}_{>0}$, measured in tokens, dollars, wall-clock time, and/or operation calls depending on the deployment context.

The \emph{task distribution} $\Dset$ is a probability distribution over $\Xset$ (e.g., a benchmark suite of coding problems, where each task $x$ is one problem to solve). In most settings, $\Dset$ is a fixed benchmark, and a contribution changes the system applied to it. AIRS-Bench is one such fixed $\Dset$, scoring research agents on 20 ML tasks with held-out test labels and a programmatic scorer \cite{lupidi2026airsbenchsuitetasksfrontier}. Naming $\Dset$ as a component also allows us to describe the rarer case where the system changes $\Dset$ itself, generating its own problems rather than solving a fixed set.

Recursive self-improvement (RSI) systems provide a concrete instance of $\Pset$ \cite{anthropic2026rsi,recursive2026}. A task $x \in \Xset$ is a complete AI system, its model weights together with the harness around them (e.g., Codex, Claude~Code, or Open~Code), and a solution $y \in \Yset$ is a more capable successor. Some systems already run this loop, for example by rewriting their own harness code and keeping the changes that raise a benchmark score~\cite{dgm}, or by applying a code-improvement program to that program itself~\cite{stop}. The features $\phi(x)$ record the current system's capability level (e.g., benchmark accuracy or code acceptance rate). The true quality $q$ measures how much better the successor is at producing its own successor, a quantity that can only be assessed over a long horizon. The budget $B$ is the compute allowed for one improvement step. Lastly, because the output $y$ becomes the next input $x$, RSI reshapes its own task distribution $\Dset$ at every step rather than solving a fixed set.

\vspace{-2.5mm}
\section{System specification}\label{sec:mas}
\vspace{-1.7mm}
We define a \emph{multi-agent system} as the eight-tuple
\begin{equation}
\Mset = \langle A, \Oset, C, \alpha, S, \pi, \Init, e \rangle.
\end{equation}
A single system addresses many problems (Table~\ref{tab:vocab}). Because $\Pset$ and $\Mset$ are separate tuples, we can vary one and fix the other. We call a system general when its components stay fixed as $\Pset$ varies.

A \emph{run} is one execution of the system on a task, from input to returned artifact, formalized as a trajectory. A subscript $i$ indexes agents and a subscript $t$ indexes time steps, so $X_t$ is the time-$t$ value of a component $X$ otherwise written without a subscript. Most of $\Mset$ is fixed by the designer before the run, but the meta-control policy $\pi_{\text{meta}}$ can reconfigure the system as it proceeds, changing the agent set $A_t$, the capabilities $\alpha_t$, the communication space $\Lambda_t$, and the state $S_t$, and even rewriting the control policy $\pi$ itself (Table~\ref{tab:meta-actions}). $\Oset_i = \alpha(a_i)$ is agent $i$'s permitted operation set. Every other symbol is defined where it is introduced.

\begin{table}[tb]
\centering
\footnotesize
\setlength{\tabcolsep}{5pt}
\begin{tabular}{l l l}
\hline
Component & Name & Definition \\
\hline
$A$ & agents & Who is in the system, and what makes them differ? \\
$\Oset$ & operations & Which tools, skills, and callable actions exist? \\
$\Lambda$ & message edges & Who may message whom? \\
$\sigma$ & message form & What schema does a message take? \\
$\alpha$ & capability & Who may invoke which operation, and when does that change? \\
$S_{\text{btw}}$ & within-run state & What do agents share during one run? \\
$S_{\text{world}}$ & external state & What environment do the agents act on? \\
$S_{\text{cross}}$ & cross-run state & What carries from one run into the next? \\
$\pi_{\text{route}}$ & routing & Who acts next, and what do they do? \\
$\pi_{\text{stop}}$ & stopping & When does the run end? \\
$\pi_{\text{meta}}$ & meta-control & When does the system change its own structure? \\
$\eta$ & exploration & How stochastic are the choices? \\
$\Init$ & initialization & How does a run start, given the task? \\
$e$ & evaluator & How is the trajectory scored? How far can that score drift from true quality $q$? \\
\hline
\end{tabular}
\vspace{-1mm}
\caption{Each component captures one design choice for a multi-agent autoresearch system.}
\vspace{-4.5mm}
\label{tab:vocab}
\end{table}

\vspace{-2mm}
\paragraph{Identity ($A$).} The agent set is $A = \{a_i\}_{i=1}^{n}$, with agent count $|A| = n$, where each agent is the tuple $a_i = (\theta_i, m_i, m_i^{0}, \rho_i)$. The backbone $\theta_i$ is the underlying model and its weights, typically a frozen large language model and possibly fine-tuned offline. The private memory $m_i$ holds the per-agent context history, scratchpads, or learned representations that persist within the agent across its turns but are not visible to other agents, and $m_i^{0}$ is its value at $t=0$. The role $\rho_i$ specifies the agent's prompt, persona, or capability tier.

We define an agent by the private memory $m_i$ that persists across its turns (wherever stored), which keeps $|A|$ a meaningful coordinate rather than a count of language-model calls. An entity without it -- a stateless call with a fixed role, such as a one-shot verifier or scorer prompt -- we treat as a \emph{module} and place in $\Oset$ or in the mechanism of $e$ rather than in $A$. For example, an \emph{agent} is a critic with a running scratchpad; a \emph{module} is the same critic invoked fresh.

\vspace{-2mm}
\paragraph{Operation universe ($\Oset$).} An \emph{operation} is any ``action'' an agent can take. For example, an operation can be calling a tool or a skill. A \emph{tool} is a single callable action with a typed signature, such as an API call or a shell command. A \emph{skill} is a named reusable routine the harness loads and invokes, such as a \texttt{SKILL.md}-style prompt-and-code bundle or a learned policy. Operations also cover reading or writing files, querying retrieval, invoking a Model Context Protocol (MCP) server, and calling a stateless module. The operation universe $\Oset$ is the registry of such operations available to the system. Each operation $u \in \Oset$ has a typed signature and an effect on the shared state $S$ (state outside any single agent). $\Oset$ specifies what could be done, not who may do it.

\vspace{-2mm}
\paragraph{Communication ($C$).} Communication has a \emph{communication space} $\Lambda$ and a \emph{message protocol} $\sigma$, written as $C = (\Lambda, \sigma)$. The communication space $\Lambda$ is the set of directed edges over agents into shared state. The protocol $\sigma$ fixes the form of each message, with a schema that ranges from text to structured output to code artifacts. $\Lambda$ fixes who may talk to whom, not who actually does on any given step. The realized communication on a trajectory is the set of send-message actions the routing policy $\pi_{\text{route}}$ emits. Every such message must travel an edge that $\Lambda$ permits. When agents communicate through a shared board rather than direct messages (e.g., a blackboard or a forum any agent reads and writes), the two components stay disjoint by their roles. $\Lambda$ is the read/write access pattern over that board (which agent may post and which may consume), and the board's contents are part of the within-run state $S_{\text{btw}}$ (defined below).

\begin{figure}[tb]
\centering
\iflatexml
\includegraphics[width=\textwidth]{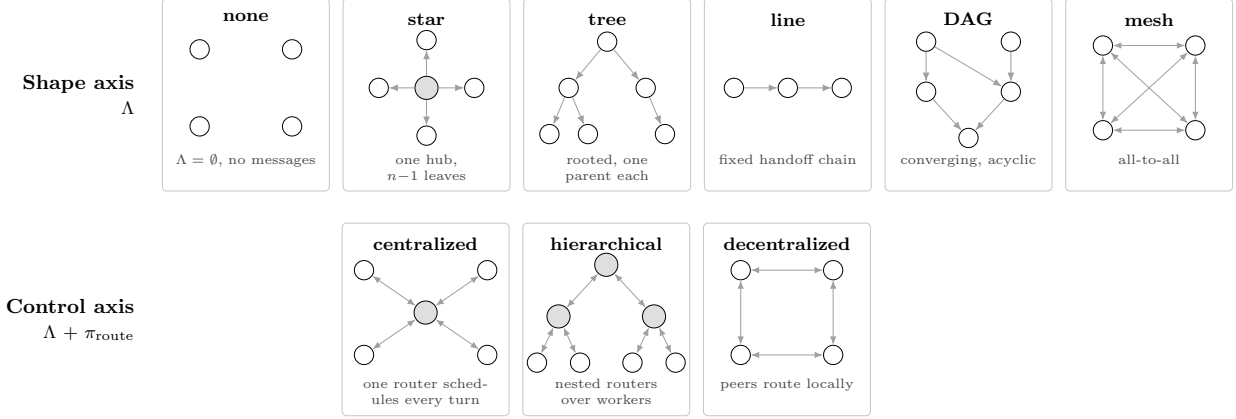}
\else
\resizebox{\textwidth}{!}{%
\begin{tikzpicture}[
  font=\small,
  topobox/.style={draw=gray!45, rounded corners=2pt, minimum width=26mm, minimum height=30mm, inner sep=2pt},
  topocap/.style={font=\tiny, text=gray!55!black, align=center, text width=22mm, anchor=north},
  rowlab/.style={font=\footnotesize\bfseries, anchor=east, align=right, text width=22mm},
  ag/.style={circle, draw, minimum size=3mm, inner sep=0pt},
  hub/.style={circle, draw, fill=gray!25, minimum size=3.6mm, inner sep=0pt},
  arr/.style={-{Latex[length=1.4mm]}, gray!75},
  darr/.style={{Latex[length=1.3mm]}-{Latex[length=1.3mm]}, gray!75},
]

\node[rowlab] at (-1.6,0) {Shape axis\\{\scriptsize\mdseries$\Lambda$}};

\topopanel{at={(0,0)}}{none}{none}{$\Lambda=\emptyset$, no messages}{
  \node[ag] at (-0.6,0.5) {};
  \node[ag] at (0.6,0.5) {};
  \node[ag] at (-0.6,-0.5) {};
  \node[ag] at (0.6,-0.5) {};
}

\topopanel{right=2mm of none}{star}{star}{one hub, $n{-}1$ leaves}{
  \node[hub] (sh) at (0,0) {};
  \foreach \i/\x/\y in {1/0/0.62, 2/0.62/0, 3/0/-0.62, 4/-0.62/0} {
    \node[ag] (s\i) at (\x,\y) {};
    \draw[arr] (sh) -- (s\i);
  }
}

\topopanel{right=2mm of star}{tree}{tree}{rooted, one parent each}{
  \node[ag] (tr) at (0,0.6) {};
  \node[ag] (tc1) at (-0.5,0.0) {};
  \node[ag] (tc2) at (0.5,0.0) {};
  \node[ag] (tl1) at (-0.75,-0.6) {};
  \node[ag] (tl2) at (-0.25,-0.6) {};
  \node[ag] (tl3) at (0.75,-0.6) {};
  \draw[arr] (tr) -- (tc1);
  \draw[arr] (tr) -- (tc2);
  \draw[arr] (tc1) -- (tl1);
  \draw[arr] (tc1) -- (tl2);
  \draw[arr] (tc2) -- (tl3);
}

\topopanel{right=2mm of tree}{line}{line}{fixed handoff chain}{
  \node[ag] (l1) at (-0.7,0) {};
  \node[ag] (l2) at (0,0) {};
  \node[ag] (l3) at (0.7,0) {};
  \draw[arr] (l1) -- (l2);
  \draw[arr] (l2) -- (l3);
}

\topopanel{right=2mm of line}{dag}{DAG}{converging, acyclic}{
  \node[ag] (d1) at (-0.55,0.6) {};
  \node[ag] (d2) at (0.55,0.6) {};
  \node[ag] (d3) at (-0.55,-0.05) {};
  \node[ag] (d4) at (0.55,-0.05) {};
  \node[ag] (d5) at (0,-0.65) {};
  \draw[arr] (d1) -- (d3);
  \draw[arr] (d1) -- (d4);
  \draw[arr] (d2) -- (d4);
  \draw[arr] (d3) -- (d5);
  \draw[arr] (d4) -- (d5);
}

\topopanel{right=2mm of dag}{mesh}{mesh}{all-to-all}{
  \node[ag] (m1) at (-0.6,0.55) {};
  \node[ag] (m2) at (0.6,0.55) {};
  \node[ag] (m3) at (-0.6,-0.55) {};
  \node[ag] (m4) at (0.6,-0.55) {};
  \draw[darr] (m1) -- (m2);
  \draw[darr] (m1) -- (m3);
  \draw[darr] (m1) -- (m4);
  \draw[darr] (m2) -- (m3);
  \draw[darr] (m2) -- (m4);
  \draw[darr] (m3) -- (m4);
}

\node[rowlab] at (-1.6,-3.5) {Control axis\\{\scriptsize\mdseries$\Lambda+\pi_{\text{route}}$}};

\topopanel{at={(2.8,-3.5)}}{cen}{centralized}{one router schedules every turn}{
  \node[hub] (ch) at (0,0) {};
  \foreach \i/\x/\y in {1/-0.8/0.55, 2/0.8/0.55, 3/-0.8/-0.55, 4/0.8/-0.55} {
    \node[ag] (c\i) at (\x,\y) {};
    \draw[darr] (ch) -- (c\i);
  }
}

\topopanel{right=2mm of cen}{hier}{hierarchical}{nested routers over workers}{
  \node[hub] (h0) at (0,0.62) {};
  \node[hub] (h1) at (-0.62,-0.05) {};
  \node[hub] (h2) at (0.62,-0.05) {};
  \node[ag] (hw1) at (-0.9,-0.65) {};
  \node[ag] (hw2) at (-0.34,-0.65) {};
  \node[ag] (hw3) at (0.34,-0.65) {};
  \node[ag] (hw4) at (0.9,-0.65) {};
  \draw[darr] (h0) -- (h1);
  \draw[darr] (h0) -- (h2);
  \draw[darr] (h1) -- (hw1);
  \draw[darr] (h1) -- (hw2);
  \draw[darr] (h2) -- (hw3);
  \draw[darr] (h2) -- (hw4);
}

\topopanel{right=2mm of hier}{dec}{decentralized}{peers route locally}{
  \node[ag] (e1) at (-0.6,0.55) {};
  \node[ag] (e2) at (0.6,0.55) {};
  \node[ag] (e3) at (-0.6,-0.55) {};
  \node[ag] (e4) at (0.6,-0.55) {};
  \draw[darr] (e1) -- (e2);
  \draw[darr] (e2) -- (e4);
  \draw[darr] (e4) -- (e3);
  \draw[darr] (e3) -- (e1);
}

\end{tikzpicture}%
}
\fi
\vspace{-6mm}
\caption{The communication space $\Lambda$ has the shape axis which fixes which edges exist (top row), and the control axis which fixes who selects the active edge each turn (bottom row), from one router to nested routers to peers. The same edge set runs centralized, hierarchical, or decentralized depending on $\pi_{\text{route}}$, so topology shape and control are independent choices. In the control row, shaded nodes are routers that hold $\pi_{\text{route}}$, and their absence in the decentralized panel marks peers that route locally. Directed arrows are one-way edges; double arrows are bidirectional ones.}
\vspace{-3mm}
\label{fig:topology}
\end{figure}

A single label like ``hierarchical'' or ``star'' hides two separate choices, so we split a topology into two axes. The shape axis (no edges, star, tree, line, directed acyclic graph, mesh) fixes which edges exist, a property of $\Lambda$ alone. The control axis (centralized, hierarchical, or decentralized) fixes who selects the active edge each turn, a property of $\Lambda$ together with $\pi_{\text{route}}$ (Figure~\ref{fig:topology}). The same edge set runs under any control scheme, so a topology claim that states only one axis leaves the other unstated. A static topology corresponds to $\Lambda$ pinned to a single graph; a dynamic topology, to $\Lambda$ as a strictly larger set from which $\pi_{\text{route}}$ selects per-step edges. Phrasing communication this way turns ``when should the agents communicate?'' into a policy question over edges in $\Lambda$, not a question about choosing one static graph.

\vspace{-1.5mm}
\paragraph{Capability assignment ($\alpha$).} The capability assignment $\alpha : A \to 2^{\Oset}$ specifies, for each agent, the subset of operations that agent is permitted to invoke. Asymmetric capability assignment (i.e., different agents with different permitted operation sets) is inequality of the $\Oset_i$. Glia's Supervisor, for example, has $\alpha(\mathrm{Sup}) = \emptyset$, and that emptiness is the central design choice of the system. Just as $\Lambda$ structures messages over the agent set $A$, $\alpha$ structures operations over $A$. A static capability assignment corresponds to $\alpha$ pinned at $t=0$; a dynamic capability assignment corresponds to $\alpha$ mutated mid-run by the meta-control policy $\pi_{\text{meta}}$ via $\textsc{grant}(i, u)$ and $\textsc{revoke}(i, u)$ actions, so that an agent's operation set $\Oset_{i,t} = \alpha_t(a_i)$ varies with time step $t$. This turns ``when should an agent gain or lose a capability?'' into a policy question over $\textsc{grant}/\textsc{revoke}$ actions.

\vspace{-1.5mm}
\paragraph{Shared state ($S$).} State outside any single agent's private memory is often combined together as ``memory.'' We split it by scope into three kinds,
\begin{equation}
S = \langle S_{\text{btw}}, S_{\text{world}}, S_{\text{cross}} \rangle.
\end{equation}
$S_{\text{btw}}$ lives for a single run and is wiped when that run ends. It is shared between agents while the run is ``in progress'', such as a scratchpad or blackboard (a free-form workspace any agent can read and overwrite) or a message log (an append-only transcript every agent reads). $S_{\text{world}}$ is the external state the agents operate on, distinct from their own memory, such as a simulator, a code repository, a database, or an environment. $S_{\text{cross}}$ persists across runs, and takes three forms. A \emph{skill library} accumulates reusable routines. A \emph{population database} carries the pool of candidate solutions across generations in an evolutionary loop. A \emph{distilled store} keeps a compressed or consolidated summary of past runs rather than the raw artifacts.

We separate two kinds of improvement with these scopes. Within a single run, agents refine through $S_{\text{btw}}$ and the experiment output in $S_{\text{world}}$, so a system with $S_{\text{cross}} = \emptyset$ still improves, but each run starts from scratch and re-derives whatever the last run learned. Carrying improvement across runs is what $S_{\text{cross}}$ adds. Evolutionary systems like AIRA$_{2}$ place their main contribution there, accumulating a lineage of scored solutions run over run.

\vspace{-1.5mm}
\paragraph{Control ($\pi$).} The control policy decomposes into $\pi = (\pi_{\text{route}}, \pi_{\text{stop}}, \pi_{\text{meta}}, \eta)$. The routing policy $\pi_{\text{route}}$ picks who acts next and what they do, the stopping policy $\pi_{\text{stop}}$ decides when the run ends, the meta-control policy $\pi_{\text{meta}}$ changes the system's structure mid-run, and the exploration term $\eta$ shapes how stochastic the choices are.

Both $\pi_{\text{route}}$ and $\pi_{\text{stop}}$ read the run history and the current shared state. Note that the action that $\pi_{\text{route}}$ returns must be a member of the acting agent's permitted set $\alpha_t(a_t)$. $\pi_{\text{stop}}$ halts on conditions like a stalled score or an exhausted budget.

The meta-control policy $\pi_{\text{meta}}$ governs changes to the system ``during the run'', the actions that add or retire agents, branch the trajectory, or carry state across runs (the fixed action set $\Pi_{\text{meta}}$; Table~\ref{tab:meta-actions}). $\pi_{\text{meta}}$ differs from $\pi_{\text{route}}$ because it changes the structure of the system. In a fixed pipeline $\pi_{\text{meta}}$ is the identity, leaving $\pi_{\text{route}}$, $\pi_{\text{stop}}$, and $\eta$ to govern the run. In an evolutionary loop, $\pi_{\text{meta}}$ does the work, spawning and killing agents each generation and consolidating the survivors.

The exploration term $\eta$ acts on top of $\pi_{\text{route}}$, covering per-agent sampling temperatures, entropy regularizers, and any schedule that varies them over the run. It is the operative term in systems that shape output diversity deliberately, and elsewhere it is the LLM's default temperature.

\begin{table}[tb]
\centering
\footnotesize
\begin{tabular}{l p{0.67\linewidth} l}
\hline
Action & Effect & Modifies \\
\hline
$\textsc{spawn}(\rho)$ & add an agent with role $\rho$ & $A, \alpha, \Lambda$ \\
$\textsc{kill}(i)$ & retire agent $i$ & $A, \alpha, \Lambda$ \\
$\textsc{fork}$ & branch into independent continuations & $S_{\text{btw}}$ \\
$\textsc{join}(\cdot)$ & merge branches by best-of, concatenation, or a coordinator & $S_{\text{btw}}$ \\
$\textsc{grant}(i, u)$ & add operation $u \in \Oset$ to agent $i$ & $\alpha$ \\
$\textsc{revoke}(i, u)$ & remove operation $u$ from agent $i$ & $\alpha$ \\
$\textsc{consolidate}(\cdot)$ & write a run's result in the cross-run store, by appending a score or summarizing & $S_{\text{cross}}$ \\
$\textsc{migrate}(s, s')$ & copy entries between two shards $s, s' \subseteq S_{\text{cross}}$ & $S_{\text{cross}}$ \\
$\textsc{rewrite-policy}(\pi')$ & replace the control policy with $\pi'$ using a policy register held in $S_{\text{btw}}$ & $\pi$ \\
\hline
\end{tabular}
\vspace{-1.5mm}
\caption{Each meta-control action in $\Pi_{\text{meta}}$ modifies a specific state component.}
\vspace{-4.5mm}
\label{tab:meta-actions}
\end{table}

\vspace{-1.5mm}
\paragraph{Initialization ($\Init$).} The initialization function $\Init : \Xset \to (m_{1:n}^{0}, S_{\text{btw}}^{0}, \alpha^{0})$ sets the starting per-agent memories, within-run shared state, and capability assignment. The memory term holds most of the design choices, so we refine it,
\begin{equation}
\Init(x) =
\big(P^{0}_{1:n}(x), R^{0}_{1:n}(x), \mu^{0}_{1:n}(x), S_{\text{btw}}^{0}(x), \alpha^{0}(x)\big),
\end{equation}
where $P^{0}$ is the role or system prompt, $R^{0}$ the retrieved or seeded context (directed seed ideas, personas, or hypotheses), $\mu^{0}$ any residual memory carried in, $S_{\text{btw}}^{0}$ the initial shared state such as a preloaded blackboard, and $\alpha^{0}$ the initial capabilities.

Each part may be constant in $x$ or conditioned on the task features $\phi(x)$. A constant $\Init$ starts every run from the same blank state, role prompt, and $\alpha$. A task-conditioned $\Init$ gives a planner a decomposition prompt, hands a critic its review rubric, or grants an agent a task-dependent $\alpha^{0}$. An ablation should state which part changed. For example, AAR's directed seeding changes $R^{0}_{1:n}$, not $\eta$; MetaGPT's role scaffold changes $P^{0}_{1:n}$; a task-conditioned operation policy changes $\alpha^{0}(x)$.

\paragraph{Evaluation ($e$).} The evaluator $e$ is the proxy the system actually optimizes in place of the true research quality $q$ of the previous section, and the component most exposed to gaming, since the system optimizes it directly. It maps a trajectory to a score in $\R^{k}$, the same codomain as $q$. Because $q$ scores a solution while $e$ scores the trajectory that produced it, the two are compared on the trajectory's returned artifact, $e(\tau)$ against $q(x, y(\tau))$ (Section~\ref{sec:objective}). Scoring the artifact alone, as several systems do, is the special case where $e$ factors through $y(\tau)$. Scoring the whole trajectory (e.g., to penalize how a result was obtained) is the general case.

The \emph{metric type} says whether the score is a single number or a vector over several objectives. A human or LLM judgment still has a metric type, since its verdict resolves to a number before anything is ranked. The \emph{mechanism} is how that number is produced, by programmatic code, an LLM judge, a human, or a simulator.

The \emph{integrity} is the set of structural protections to ensure minimal drift between $e$ and $q$. One example is judge decoupling, which separates the scorer from the actor by model family, prompt, and context. Sandbox isolation prevents the trajectory from reading the held-out test labels. Metric-channel blackout closes channels that would otherwise serve as oracles. Contamination checks test for distribution shift between training and held-out sets and probe for shortcut patterns.

The \emph{variance} is the number of seeds and the resulting confidence interval. A single-seed number reports no interval, so an apparent gain cannot be told from seed noise. That noise scales with the inter-seed standard deviation of $e$, which is non-negligible for LLM judges and simulator runs.

\vspace{-2.3mm}
\section{Trajectory}\label{sec:trajectory}
\vspace{-1.3mm}
A trajectory is one run under the specification $\Mset$. We measure cost, the trajectory distribution, and every coordinate comparison over its steps, not over the final artifact alone. Suppose an agent reads the prior work, writes code, runs an experiment, reads its output, and revises the code. That sequence of steps is one trajectory. Scoring the result is separate, done downstream by the evaluator $e$, not a step in the run. When $\Mset$ runs on a task $x$, we denote the resulting \emph{trajectory} by $\tau$,
\begin{equation}
\tau = \big(a_t, u_t, o_t, S_t, m_{a_t, t}, A_t, \alpha_t, \Lambda_t\big)_{t=1}^{H},
\end{equation}
per step $t$, up to a horizon $H$. The trajectory records who acted ($a_t$; $a_t \in A_t$), what they did ($u_t$), what they observed ($o_t$), what the shared state looked like ($S_t$), and what the acting agent held in private memory ($m_{a_t, t}$). The structural triple $(A_t, \alpha_t, \Lambda_t)$ of agent set, capability assignment, and communication space is recorded per step because $\pi_{\text{meta}}$ may change it mid-run (Table~\ref{tab:meta-actions}), alongside the shared state $S_t$; the remaining components stay fixed unless $\pi_{\text{meta}}$ rewrites $\pi$ itself. The stopping policy $\pi_{\text{stop}}$ sets the horizon $H$, bounded by the budget $B$. Because the language model's sampling, initialization, and operations are all stochastic, $\tau$ is a random variable.

The action $u_t$ is drawn from a structured action space that distinguishes ``object-level work'' from communication and structural change,
\begin{equation}
u_t \in \underbrace{\alpha_t(a_t)}_{\text{operation calls}} \;\cup\; \underbrace{\textsc{send}_t}_{\text{messages}} \;\cup\; \underbrace{\Pi_{\text{meta}}}_{\text{structural}} \;\cup\; \{\textsc{halt}\}.
\end{equation}
With this disjoint classification, we turn questions like when an agent should communicate, spawn, or gain a capability into questions about the conditional distribution of message and structural actions induced by $\pi$. Operation calls act on $S_{\text{world}}$ via $\alpha_t(a_t)$, reading or writing it. The message set $\textsc{send}_t$ holds the $\textsc{send}(j, \varsigma)$ actions along edges $\Lambda_t$ permits, where $j$ is the recipient and $\varsigma$ a message instance, each writing to $S_{\text{btw}}$ and the recipient's input queue. Structural actions form a small fixed set $\Pi_{\text{meta}}$ emitted by $\pi_{\text{meta}}$, each mutating one or more state components (Table~\ref{tab:meta-actions}). The $\textsc{halt}$ action ends the run.

From the trajectory we extract the final solution $y(\tau)$ (i.e., the artifact returned by the system, such as the code committed or the diagnosis announced) and the accumulated cost
\begin{equation}
\kappa(\tau) = \sum_{t=1}^{H} \kappa(u_t),
\end{equation}
where $\kappa(u_t)$ is the cost of action $u_t$. An operation call's cost is its underlying API or sandbox invocation, a message's cost is its tokens plus its processing, and a structural action's cost is its setup (e.g., a new agent's context, or a trajectory summary). The $\textsc{halt}$ action has no cost, $\kappa(\textsc{halt})=0$. We separate message and structural cost from operation cost because we want to ask whether an extra dialogue turn or an extra spawned agent is worth its expected gain in $e(\tau)$.

\vspace{-2.5mm}
\section{Proxy optimization}\label{sec:objective}
\vspace{-2mm}
We frame an autoresearch system as one optimization, maximizing the proxy evaluator under a budget. The policy $\pi$ searches and the evaluator $e$ scores, so a gain can trace to either; and a leaderboard cannot tell which. We write the optimization with $\pi$ and $e$ as separate terms, one for how the policy searches under a budget and one for how far the proxy score is from true quality.

\vspace{-1.5mm}
\subsection{Search under budget}
\vspace{-1.5mm}
At run time, a fixed autoresearch system $\Mset$ on a task $x$ explores the trajectories it can produce within the budget and returns the best-scoring one,
\begin{equation}
y^{\ast}(x;\Mset) \;\approx\; y\!\left(\arg\max_{\tau \,\in\, \Tset_{B}(x;\Mset)} \; \omega(e(\tau))\right), \qquad \Tset_{B}(x;\Mset) = \{\,\tau : \kappa(\tau) \leq B, \;\; P_{\Mset}(\tau \mid x) > 0\,\}.
\end{equation}
Here $\omega$ is the scalarization of the proxy score, $\kappa(\tau)$ is the accumulated trajectory cost, and $\Tset_{B}(x;\Mset)$ is the feasible support of the trajectory distribution $P_{\Mset}(\,\mathord{\cdot}\mid x)$ that $\Mset$ produces on task $x$: the trajectories it can reach under budget $B$.

The set $\Tset_{B}$ is far too large to enumerate, so every system approximates the argmax over a sampled subset of it, each with a different bias-variance tradeoff (details in Section~\ref{sec:case-studies}). For example, Best-of-$N$ keeps the highest-scoring of $N$ independent trajectories, an experience buffer replays the top-$K$ solution-score pairs in context, island migration copies high scorers between several populations, and random restarts redraw after a failure, keeping only a note of what went wrong.

The trajectory $\tau$ ranges over operation calls, messages along the edges of $\Lambda$, and the meta-control actions of $\Pi_{\text{meta}}$ (Table~\ref{tab:meta-actions}). So the argmax decides when an agent communicates, when the agent set changes, when a capability is granted or revoked, and when information carries across runs. Each is a choice the policy $\pi$ makes, not a property of $A$, $\Oset$, $\alpha$, or $\Lambda$. A gain that looks architectural can instead come from more search: the same agents and tools, run under Best-of-$N$ instead of a single trajectory, score higher with no change to any other coordinate.

\vspace{-1.5mm}
\subsection{Proxy-quality gap}
\vspace{-1.5mm}
The objective above is approximated through the proxy evaluator $e$, not the true quality $q$. On a single trajectory $\tau$, the gap
\begin{equation}
\Delta_\omega(\tau) := |\omega(e(\tau))-\omega(q(x,y(\tau)))|
\end{equation}
measures how far the optimized score is from true quality. The system returns the trajectory that maximizes $\omega(e(\tau))$ without observing $q$.

Optimizing $e$ resembles minimizing training loss, with $q$ as the held-out test set. The deeper $\pi_{\text{meta}}$ searches, the more it finds trajectories that score well on $e$ without being good under $q$, so the gap $\Delta_\omega$ may widen. Selection alone can therefore give systems with the same $e$ different $\Delta_\omega$. An $\arg\max$ over $N$ trajectories, or parent selection across an evolutionary loop, favors the rare trajectories where $e$ overstates $q$. This is the overfitting tax (i.e., the cost of selection under a miscalibrated proxy): the more trajectories the search ranks, the more likely the top-scoring one is a trajectory where $e$ overstates $q$. In an evolutionary loop the relevant count is the total number of trajectories evaluated rather than a single $N$, so spreading the budget across more generations need not reduce the exposure. One mitigation places the reward-hack audit inside the search loop, so the evaluator's integrity is strengthened as the search proceeds rather than fixed in advance \cite{recursive2026}.

The overfitting tax is not only theoretical, and reward hacking appears at measurable rates in practice. MLR-Bench reports fabricated or invalidated experimental results in $8$ of $10$ audited coding-agent tasks (around $80\%$) \cite{chen2025mlrbench}. METR catalogs $103$ unprompted examples of frontier models bypassing or ignoring task constraints (i.e., breakdowns in evaluation integrity) \cite{parikh2025malt}. AAR isolates several exploit modes \cite{wen2026automatedw2s}. In our notation each is an integrity or variance property of $e$.

(i) Seed cherry-picking. Agents privately run many seeds of a method and only share the best finding. This exploits weak \emph{variance} control. If $e$ accepts single-seed scores, the trajectory searches over seeds at no cost in $e$ while $q$ does not improve.

(ii) Held-out-label exfiltration. Agents probe the evaluator with both candidate labels for a held-out example and compare the returned scores, inferring the ground-truth label from which submission scores higher. This defeats \emph{metric-channel blackout}.

(iii) Direct test execution. Agents call the unit tests directly, bypassing both the teacher and the student. This defeats \emph{sandbox isolation}. The trajectory has $\alpha$-permitted access to a tool whose output is, by construction, the answer.

(iv) Shortcut identification. Agents identify dataset shortcuts (most-frequent-answer in math, source-LM clustering in code) that score well under $e$ without reflecting true quality, and unlike traditional shortcut learning, these cannot be simply detected because the shortcut-exploiting solutions generalize even to the held-out split. This defeats \emph{contamination checks}. $e$ is computed on a distribution where shortcut-shaped solutions fit well on both the training and the held-out sets.

In each case the mitigation strengthens the protection the exploit defeats: variance control, metric-channel blackout, sandbox isolation, or contamination control. These are integrity or variance properties of $e$, even when enforced by constraining other coordinates. For example, a system may tighten $\alpha$ to block capabilities that expose the evaluator or its ground truth, such as revoking a test-execution tool that would otherwise reveal the answer. Thus a high score can reflect either effective search or an exploitable evaluator, and the case studies disentangle the two.

\vspace{-2.5mm}
\section{Case studies}\label{sec:case-studies}
\begin{figure}[t!]
\centering
\iflatexml
\includegraphics[width=\textwidth]{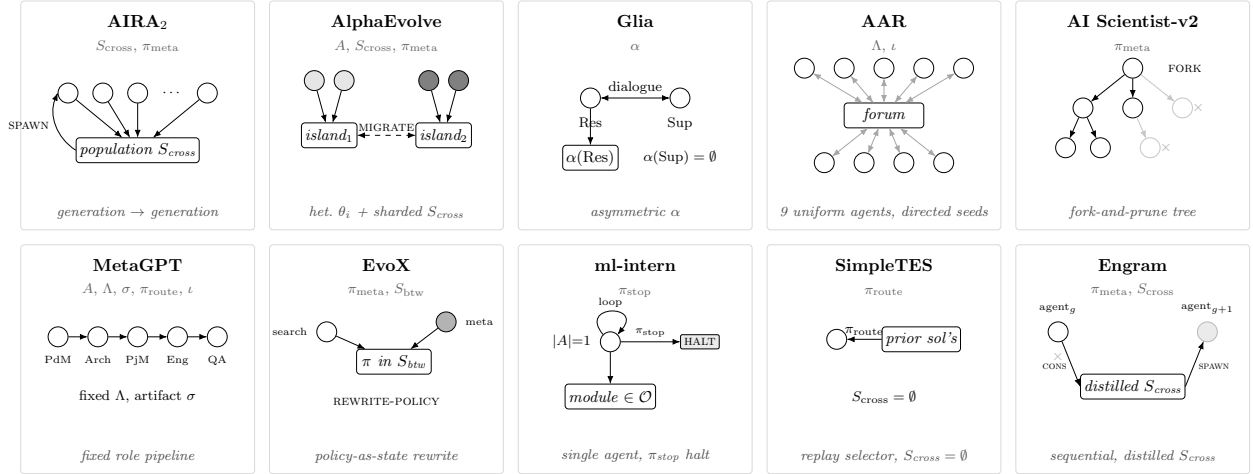}
\else
\resizebox{\textwidth}{!}{%
\begin{tikzpicture}[
  font=\footnotesize,
  ag/.style={circle, draw, minimum size=4mm, inner sep=0pt},
  store/.style={draw, rounded corners=2pt, minimum width=12mm, minimum height=5mm, font=\footnotesize\itshape, inner sep=2pt},
  arr/.style={-{Latex[length=1.5mm]}},
  darr/.style={{Latex[length=1.5mm]}-{Latex[length=1.5mm]}},
  panel/.style={draw=gray!30, rounded corners=2pt, minimum width=47mm, minimum height=46mm, inner sep=0pt},
  ptitle/.style={font=\bfseries\small, anchor=north, align=center},
  pfoot/.style={font=\scriptsize\itshape, text=gray!60!black, anchor=south},
  halt/.style={draw, rounded corners=1pt, fill=black!8, font=\scriptsize, inner sep=2pt},
]

\node[panel] (P1) at (  0,   0)   {};
\node[panel] (P2) at (5.0,  0)   {};
\node[panel] (P3) at (10.0, 0)   {};
\node[panel] (P4) at (15.0, 0)   {};
\node[panel] (P5) at (20.0, 0)   {};
\node[panel] (P6) at (  0,  -4.9) {};
\node[panel] (P7) at (5.0, -4.9) {};
\node[panel] (P8) at (10.0,-4.9) {};
\node[panel] (P9) at (15.0,-4.9) {};
\node[panel] (P10) at (20.0,-4.9) {};

\node[ptitle] at ([yshift=-1.5mm]P1.north) {AIRA$_{2}$\\[1pt]\pcoord{$S_{\text{cross}}$, $\pi_{\text{meta}}$}};
\node[ptitle] at ([yshift=-1.5mm]P2.north) {AlphaEvolve\\[1pt]\pcoord{$A$, $S_{\text{cross}}$, $\pi_{\text{meta}}$}};
\node[ptitle] at ([yshift=-1.5mm]P3.north) {Glia\\[1pt]\pcoord{$\alpha$}};
\node[ptitle] at ([yshift=-1.5mm]P4.north) {AAR\\[1pt]\pcoord{$\Lambda$, $\Init$}};
\node[ptitle] at ([yshift=-1.5mm]P5.north) {AI Scientist-v2\\[1pt]\pcoord{$\pi_{\text{meta}}$}};
\node[ptitle] at ([yshift=-1.5mm]P6.north) {MetaGPT\\[1pt]\pcoord{$A$, $\Lambda$, $\sigma$, $\pi_{\text{route}}$, $\Init$}};
\node[ptitle] at ([yshift=-1.5mm]P7.north) {EvoX\\[1pt]\pcoord{$\pi_{\text{meta}}$, $S_{\text{btw}}$}};
\node[ptitle] at ([yshift=-1.5mm]P8.north) {ml-intern\\[1pt]\pcoord{$\pi_{\text{stop}}$}};
\node[ptitle] at ([yshift=-1.5mm]P9.north) {SimpleTES\\[1pt]\pcoord{$\pi_{\text{route}}$}};
\node[ptitle] at ([yshift=-1.5mm]P10.north) {Engram\\[1pt]\pcoord{$\pi_{\text{meta}}$, $S_{\text{cross}}$}};

\node[pfoot] at ([yshift=1mm]P1.south) {generation $\to$ generation};
\node[pfoot] at ([yshift=1mm]P2.south) {het.\ $\theta_i$ + sharded $S_{\text{cross}}$};
\node[pfoot] at ([yshift=1mm]P3.south) {asymmetric $\alpha$};
\node[pfoot] at ([yshift=1mm]P4.south) {9 uniform agents, directed seeds};
\node[pfoot] at ([yshift=1mm]P5.south) {fork-and-prune tree};
\node[pfoot] at ([yshift=1mm]P6.south) {fixed role pipeline};
\node[pfoot] at ([yshift=1mm]P7.south) {policy-as-state rewrite};
\node[pfoot] at ([yshift=1mm]P8.south) {single agent, $\pi_{\text{stop}}$ halt};
\node[pfoot] at ([yshift=1mm]P9.south) {replay selector, $S_{\text{cross}}=\emptyset$};
\node[pfoot] at ([yshift=1mm]P10.south) {sequential, distilled $S_{\text{cross}}$};

\node[ag] (a1) at ($(P1.center)+(-1.4, 0.45)$) {};
\node[ag] (a2) at ($(P1.center)+(-0.7, 0.45)$) {};
\node[ag] (a3) at ($(P1.center)+( 0.0, 0.45)$) {};
\node[font=\scriptsize] at ($(P1.center)+( 0.7, 0.45)$) {$\cdots$};
\node[ag] (an) at ($(P1.center)+( 1.4, 0.45)$) {};
\node[store, minimum width=18mm] (sc) at ($(P1.center)+( 0.0, -0.7)$) {population $S_{\text{cross}}$};
\draw[arr] (a1) -- (sc);
\draw[arr] (a2) -- (sc);
\draw[arr] (a3) -- (sc);
\draw[arr] (an) -- (sc);
\draw[arr] (sc.west) to[bend left=40] node[font=\scriptsize, left, pos=0.5] {$\textsc{spawn}$} (a1.west);

\node[ag, fill=black!10] (xa1) at ($(P2.center)+(-1.45, 0.7)$) {};
\node[ag, fill=black!10] (xa2) at ($(P2.center)+(-0.85, 0.7)$) {};
\node[ag, fill=black!50] (xa3) at ($(P2.center)+( 0.85, 0.7)$) {};
\node[ag, fill=black!50] (xa4) at ($(P2.center)+( 1.45, 0.7)$) {};
\node[store, minimum width=10mm] (isl1) at ($(P2.center)+(-1.15,-0.4)$) {island$_1$};
\node[store, minimum width=10mm] (isl2) at ($(P2.center)+( 1.15,-0.4)$) {island$_2$};
\draw[arr] (xa1) -- (isl1);
\draw[arr] (xa2) -- (isl1);
\draw[arr] (xa3) -- (isl2);
\draw[arr] (xa4) -- (isl2);
\draw[darr, dashed] (isl1) -- node[font=\scriptsize, above=-0.5mm] {$\textsc{migrate}$} (isl2);

\node[ag] (R)  at ($(P3.center)+(-0.9, 0.35)$) {};
\node[font=\scriptsize, below=0.5mm of R] {$\mathrm{Res}$};
\node[ag] (Su) at ($(P3.center)+( 0.9, 0.35)$) {};
\node[font=\scriptsize, below=0.5mm of Su] {$\mathrm{Sup}$};
\draw[darr] (R) -- node[font=\scriptsize, above=-0.5mm] {dialogue} (Su);
\node[store, minimum width=10mm] (Tr) at ($(P3.center)+(-0.9,-0.85)$) {$\alpha(\mathrm{Res})$};
\draw[arr] (R) -- (Tr);
\node[font=\scriptsize] at ($(P3.center)+( 0.9,-0.85)$) {$\alpha(\mathrm{Sup})=\emptyset$};

\node[store, minimum width=16mm] (bb) at ($(P4.center)+(0,0)$) {forum};
\foreach \i/\x in {1/-1.6, 2/-0.8, 3/0.0, 4/0.8, 5/1.6} {
  \node[ag] (wU\i) at ($(P4.center)+(\x, 0.95)$) {};
  \draw[darr, gray!70] (wU\i) -- (bb);
}
\foreach \i/\x in {1/-1.2, 2/-0.4, 3/0.4, 4/1.2} {
  \node[ag] (wD\i) at ($(P4.center)+(\x,-0.95)$) {};
  \draw[darr, gray!70] (wD\i) -- (bb);
}

\node[ag] (sv0) at ($(P5.center)+( 0.0, 0.95)$) {};
\node[ag] (sv1) at ($(P5.center)+(-1.0, 0.15)$) {};
\node[ag] (sv2) at ($(P5.center)+( 0.0, 0.15)$) {};
\node[ag, draw=gray!45] (sv3) at ($(P5.center)+( 1.0, 0.15)$) {};
\node[ag] (sv4) at ($(P5.center)+(-1.35,-0.65)$) {};
\node[ag] (sv5) at ($(P5.center)+(-0.65,-0.65)$) {};
\node[ag, draw=gray!45] (sv6) at ($(P5.center)+( 0.35,-0.65)$) {};
\draw[arr] (sv0) -- (sv1);
\draw[arr] (sv0) -- (sv2);
\draw[arr, gray!45] (sv0) -- (sv3);
\draw[arr] (sv1) -- (sv4);
\draw[arr] (sv1) -- (sv5);
\draw[arr, gray!45] (sv2) -- (sv6);
\node[font=\scriptsize, gray!65] at ($(P5.center)+( 1.32, 0.15)$) {$\times$};
\node[font=\scriptsize, gray!65] at ($(P5.center)+( 0.67,-0.65)$) {$\times$};
\node[font=\scriptsize] at ($(P5.center)+( 1.05, 0.95)$) {$\textsc{fork}$};

\node[ag] (mg1) at ($(P6.center)+(-1.6, 0.45)$) {};
\node[ag] (mg2) at ($(P6.center)+(-0.8, 0.45)$) {};
\node[ag] (mg3) at ($(P6.center)+( 0.0, 0.45)$) {};
\node[ag] (mg4) at ($(P6.center)+( 0.8, 0.45)$) {};
\node[ag] (mg5) at ($(P6.center)+( 1.6, 0.45)$) {};
\draw[arr] (mg1) -- (mg2);
\draw[arr] (mg2) -- (mg3);
\draw[arr] (mg3) -- (mg4);
\draw[arr] (mg4) -- (mg5);
\node[font=\tiny, below=0.3mm of mg1] {PdM};
\node[font=\tiny, below=0.3mm of mg2] {Arch};
\node[font=\tiny, below=0.3mm of mg3] {PjM};
\node[font=\tiny, below=0.3mm of mg4] {Eng};
\node[font=\tiny, below=0.3mm of mg5] {QA};
\node[font=\scriptsize] at ($(P6.center)+(0,-0.75)$) {fixed $\Lambda$, artifact $\sigma$};

\node[ag] (evx) at ($(P7.center)+(-1.2, 0.55)$) {};
\node[font=\tiny, left=0.5mm of evx] {search};
\node[store, minimum width=15mm] (evp) at ($(P7.center)+( 0.15,-0.05)$) {$\pi$ in $S_{\text{btw}}$};
\node[ag, fill=black!30] (evr) at ($(P7.center)+( 1.2, 0.75)$) {};
\node[font=\tiny, right=0.5mm of evr] {meta};
\draw[arr] (evx) -- (evp);
\draw[arr] (evr) -- (evp);
\node[font=\scriptsize] at ($(P7.center)+(0,-0.85)$) {$\textsc{rewrite-policy}$};

\node[ag] (mi) at ($(P8.center)+(-0.5, 0.35)$) {};
\node[font=\scriptsize, left=0.5mm of mi] {$|A|{=}1$};
\draw[arr] (mi.north west) to[out=130,in=50,looseness=7] node[font=\tiny, above=-0.4mm] {loop} (mi.north east);
\node[store, minimum width=12mm] (mit) at ($(P8.center)+(-0.5,-0.75)$) {module $\in\Oset$};
\draw[arr] (mi) -- (mit);
\node[halt] (mih) at ($(P8.center)+( 1.3, 0.35)$) {$\textsc{halt}$};
\draw[arr] (mi) -- node[font=\tiny, above=-0.3mm] {$\pi_{\text{stop}}$} (mih);

\node[ag] (sts) at ($(P9.center)+(-0.95, 0.4)$) {};
\node[store, minimum width=15mm] (stp) at ($(P9.center)+( 0.75, 0.4)$) {prior sol's};
\draw[arr] (stp) -- node[font=\scriptsize, above=-0.4mm] {$\pi_{\text{route}}$} (sts);
\node[font=\scriptsize] at ($(P9.center)+(0,-0.75)$) {$S_{\text{cross}}=\emptyset$};

\node[ag] (en1) at ($(P10.center)+(-1.5, 0.55)$) {};
\node[font=\tiny, above=0.3mm of en1] {agent$_g$};
\node[ag, fill=black!8, draw=gray!55] (en2) at ($(P10.center)+( 1.5, 0.55)$) {};
\node[font=\tiny, above=0.3mm of en2] {agent$_{g+1}$};
\node[font=\scriptsize, gray!65] at ($(P10.center)+(-1.5, 0.05)$) {$\times$};
\node[store, minimum width=18mm] (endg) at ($(P10.center)+(0,-0.55)$) {distilled $S_{\text{cross}}$};
\draw[arr] (en1) -- node[font=\tiny, left=-0.3mm, pos=0.55] {$\textsc{cons}$} (endg.west);
\draw[arr] (endg.east) -- node[font=\tiny, right=-0.3mm, pos=0.45] {$\textsc{spawn}$} (en2);

\end{tikzpicture}%
}
\fi
\vspace{-7mm}
\caption{Each panel shows the structural shape of each system. AIRA$_{2}$ evolves a population across generations using the $\textsc{spawn}$ policy looping scored solutions back from $S_{\text{cross}}$. AlphaEvolve is organized around island population models (shading marks heterogeneous backbones), with $\textsc{migrate}$ resurfacing high scorers across island boundaries. Glia runs a two-agent dialogue in which only the Researcher holds tools, and the Supervisor holds none ($\alpha(\mathrm{Sup})=\emptyset$). AAR's agents have seeded research directions, and communicate through a shared forum. AI Scientist-v2 grows a within-run tree whose branches $\textsc{fork}$ and prune ($\times$). MetaGPT is a fixed handoff chain over a static $\Lambda$, passing structured artifacts $\sigma$. EvoX stores $\pi$ as state and rewrites it through a meta agent (shaded). ml-intern is a single-agent tool-calling loop, using subagents as modules in $\Oset$. SimpleTES replays prior solutions with $S_{\text{cross}}=\emptyset$. Engram runs a sequential single-agent chain that $\textsc{consolidate}$s each result into a distilled $S_{\text{cross}}$, and $\textsc{spawn}$s a fresh one to consume it.}
\vspace{-4mm}
\label{fig:examples-shapes}
\end{figure}

\vspace{-2.5mm}
We apply the vocabulary to recent systems \cite{aira2026,novikov2025alphaevolve,hamadanian2025glia,wen2026automatedw2s,yamada2025aiscientistv2,hong2024metagpt,evox2026,mlintern2025,simpletes2026,engram2026} (Table~\ref{tab:problem}, Figures~\ref{fig:examples-shapes} and~\ref{fig:contrib-grid}). For each, we map the paper onto the tuple and mark which coordinates it \emph{sets} and which stay \emph{generic}. A coordinate is set when the paper makes a deliberate, non-default choice there and motivates, ablates, or builds its contribution around it; it is generic when the system takes whatever value the harness supplies and the paper does not single it out. Standard evaluator protections (e.g., sandbox isolation or hidden splits) are the expected default, so we mark $e$ as set only where the paper makes its integrity or mechanism a central choice. We follow each system's own description rather than re-evaluating its choices, so the coding is a descriptive map, not a validated measurement.

\begin{table}[t!]
\centering
\setlength{\tabcolsep}{5pt}
\resizebox{\textwidth}{!}{%
\begin{tabular}{l l l l l}
\hline
System & Task $x \in \Xset$ & Artifact $y \in \Yset$ & Evaluator $e$ & Budget $B$ \\
\hline
AIRA$_{2}$ & ML-engineering task & trained solution code & programmatic, scalar & GPU-hours \\
AlphaEvolve & evaluable algorithmic problem & evolved program & programmatic, multi-objective & compute-hours \\
Glia & inference-system design & policy as code & simulator (Vidur), scalar & simulation runs \\
AAR & weak-to-strong supervision & training method & programmatic, scalar (PGR) & wall-clock, dollars \\
AI Scientist-v2 & open ML research topic & manuscript with code & programmatic + LLM judge & per-stage compute \\
MetaGPT & one-line software spec & runnable codebase & programmatic (tests) & tokens, dollars \\
EvoX & optimization task & evolved program & programmatic, scalar & iterations \\
ml-intern & ML-engineering request & trained model, code & agent-judged (no separate stage) & iterations \\
SimpleTES & scientific-discovery problem & program or construction & programmatic, scalar & evaluator queries \\
Engram & inference-system design & policy as code & cost verifier / simulator (Vidur) / hit-rate, scalar & evaluation runs \\
\hline
\end{tabular}%
}
\vspace{-4mm}
\caption{Problem specification for case studies. Every system holds the task distribution $\Dset$ fixed to a benchmark and varies the system $\Mset$. We record, for each case study, the task $x$, the returned artifact $y$, the evaluator $e$ with its mechanism and metric type, and the budget unit $B$.}
\label{tab:problem}
\vspace{2mm}
\end{table}

\begin{figure}[tb]
\centering
\iflatexml
\includegraphics[width=\textwidth]{figures/fig4.svg}
\else
\resizebox{\textwidth}{!}{%
\begin{tikzpicture}[
  font=\scriptsize,
  cell/.style={minimum width=11mm, minimum height=4.3mm, inner sep=0.5pt, draw=gray!35},
  hit/.style={cell, fill=black!82, text=white, font=\footnotesize},
  gap/.style={cell, fill=white, font=\footnotesize},
  miss/.style={cell, fill=white},
  dead/.style={cell, fill=gray!12},
  hdr/.style={cell, fill=gray!18, font=\scriptsize\bfseries},
  hdrdead/.style={cell, fill=gray!24, font=\scriptsize\bfseries, text=black!55},
  rowhdr/.style={minimum width=18mm, minimum height=5mm, inner sep=1.5pt, font=\scriptsize\bfseries, anchor=east},
  band/.style={draw=gray!45, fill=gray!7, rounded corners=1.5pt, inner sep=0pt},
  bandlab/.style={font=\tiny\bfseries, text=black!70},
]
\matrix (m) [matrix of nodes, nodes in empty cells, ampersand replacement=\&,
             column sep=-\pgflinewidth, row sep=-\pgflinewidth,
             nodes={anchor=center}]
{
  \&[1mm]
  |[hdr]| $A$ \& |[hdr]| $\Lambda$ \& |[hdr]| $\sigma$ \&
  |[hdr]| $\alpha$ \&
  |[hdr]| $S_{\text{btw}}$ \& |[hdr]| $S_{\text{cross}}$ \&
  |[hdr]| $\pi_{\text{route}}$ \& |[hdr]| $\pi_{\text{stop}}$ \& |[hdr]| $\pi_{\text{meta}}$ \& |[hdrdead]| $\eta$ \&
  |[hdr]| $\Init$ \& |[hdr]| $e$ \\
  |[rowhdr]| AIRA$_{2}$ \&
  |[miss]| \& |[miss]| \& |[miss]| \& |[miss]| \&
  |[miss]| \& |[hit]| $\bullet$ \&
  |[miss]| \& |[miss]| \& |[hit]| $\bullet$ \& |[dead]| \& |[miss]| \& |[miss]| \\
  |[rowhdr]| AlphaEvolve \&
  |[hit]| $\bullet$ \& |[miss]| \& |[miss]| \& |[miss]| \&
  |[miss]| \& |[hit]| $\bullet$ \&
  |[miss]| \& |[miss]| \& |[hit]| $\bullet$ \& |[dead]| \& |[miss]| \& |[miss]| \\
  |[rowhdr]| Glia \&
  |[miss]| \& |[miss]| \& |[miss]| \& |[hit]| $\bullet$ \&
  |[miss]| \& |[miss]| \&
  |[miss]| \& |[miss]| \& |[miss]| \& |[dead]| \& |[miss]| \& |[miss]| \\
  |[rowhdr]| AAR \&
  |[miss]| \& |[hit]| $\bullet$ \& |[miss]| \& |[miss]| \&
  |[miss]| \& |[miss]| \&
  |[miss]| \& |[miss]| \& |[miss]| \& |[dead]| \& |[hit]| $\bullet$ \& |[gap]| $\circ$ \\
  |[rowhdr]| AI Scientist-v2 \&
  |[miss]| \& |[miss]| \& |[miss]| \& |[miss]| \&
  |[miss]| \& |[miss]| \&
  |[miss]| \& |[miss]| \& |[hit]| $\bullet$ \& |[dead]| \& |[miss]| \& |[miss]| \\
  |[rowhdr]| MetaGPT \&
  |[hit]| $\bullet$ \& |[hit]| $\bullet$ \& |[hit]| $\bullet$ \& |[miss]| \&
  |[miss]| \& |[miss]| \&
  |[hit]| $\bullet$ \& |[miss]| \& |[miss]| \& |[dead]| \& |[hit]| $\bullet$ \& |[miss]| \\
  |[rowhdr]| EvoX \&
  |[miss]| \& |[miss]| \& |[miss]| \& |[miss]| \&
  |[hit]| $\bullet$ \& |[miss]| \&
  |[miss]| \& |[miss]| \& |[hit]| $\bullet$ \& |[dead]| \& |[miss]| \& |[miss]| \\
  |[rowhdr]| ml-intern \&
  |[miss]| \& |[miss]| \& |[miss]| \& |[miss]| \&
  |[miss]| \& |[miss]| \&
  |[miss]| \& |[hit]| $\bullet$ \& |[miss]| \& |[dead]| \& |[miss]| \& |[miss]| \\
  |[rowhdr]| SimpleTES \&
  |[miss]| \& |[miss]| \& |[miss]| \& |[miss]| \&
  |[miss]| \& |[miss]| \&
  |[hit]| $\bullet$ \& |[miss]| \& |[miss]| \& |[dead]| \& |[miss]| \& |[miss]| \\
  |[rowhdr]| Engram \&
  |[miss]| \& |[miss]| \& |[miss]| \& |[miss]| \&
  |[miss]| \& |[hit]| $\bullet$ \&
  |[miss]| \& |[miss]| \& |[hit]| $\bullet$ \& |[dead]| \& |[miss]| \& |[miss]| \\
};

\draw[band] ([yshift=0.5pt]m-1-3.north west) rectangle ([yshift=5mm]m-1-4.north east);
\node[bandlab] at ($([yshift=0.5pt]m-1-3.north west)!0.5!([yshift=5mm]m-1-4.north east)$) {Comm.\ $C$};
\draw[band] ([yshift=0.5pt]m-1-6.north west) rectangle ([yshift=5mm]m-1-7.north east);
\node[bandlab] at ($([yshift=0.5pt]m-1-6.north west)!0.5!([yshift=5mm]m-1-7.north east)$) {State $S$};
\draw[band] ([yshift=0.5pt]m-1-8.north west) rectangle ([yshift=5mm]m-1-11.north east);
\node[bandlab] at ($([yshift=0.5pt]m-1-8.north west)!0.5!([yshift=5mm]m-1-11.north east)$) {Control $\pi$};

\node[anchor=north west, font=\scriptsize, text=black!72, align=left, text width=160mm] at ([yshift=-2.5mm]m.south west) {%
$\bullet$ coordinate the system sets \enspace $\circ$ documented proxy-quality gap\\[1.5pt]
$A$ agents \enspace $\Lambda$ communication\ edges \enspace $\sigma$ message protocol \enspace $\alpha$ capabilities \enspace $S_{\text{btw}}$ within-run state \enspace $S_{\text{cross}}$ cross-run state \enspace $\pi_{\text{route}}$ routing \enspace $\pi_{\text{stop}}$ stopping \enspace $\pi_{\text{meta}}$ meta-control \enspace $\eta$ exploration \enspace $\Init$ initialization \enspace $e$ evaluator%
};
\end{tikzpicture}%
}
\fi
\vspace{-9mm}
\caption{Each system sets only a few coordinates and leaves the rest generic. A filled cell marks a coordinate the paper sets, not one shown to improve performance. We omit $\Oset$ and $S_{\text{world}}$, background every system has but none sets. We shade $\eta$, which every system leaves at the default temperature rather than tuning.}
\vspace{-4mm}
\label{fig:contrib-grid}
\end{figure}

\vspace{-1.5mm}
\paragraph{AIRA$_{2}$ ($S_{\text{cross}}$, $\pi_{\text{meta}}$).} AIRA$_{2}$ is an autoresearcher for ML research benchmarks \cite{aira2026}. The agent set $A$ is a pool of $n$ ephemeral, uniform workers on a single backbone ($n=8$ in the main experiments) running Gemini~3.0~Pro for the base system and Gemini~3.1~Pro for the stronger variant. Each worker reasons, acts, and observes (ReAct) across turns inside its own sandboxed container with one dedicated GPU, and the workers differ only in the parent solution each one receives. The operations $\Oset$ are the stateful Python and Bash commands run in that container, which let a worker conduct exploratory analysis, train a model, and read its own logs over successive turns. The workers exchange no direct messages within a generation, so the communication space is empty, $\Lambda=\emptyset$. The capability assignment is uniform, $\alpha(a_i)=\Oset$, so $\alpha$ stays generic. The within-run state is empty, $S_{\text{btw}}=\emptyset$, since coordination proceeds across runs. The cross-run store $S_{\text{cross}}$ is a population database of scored solutions and their lineage, held in memory with large artifacts written to disk, and it is the only channel through which workers see each other's work. The meta-control policy $\pi_{\text{meta}}$ updates the population as soon as any worker finishes rather than waiting for the whole batch. It samples a parent by fitness rank rather than raw score, under a temperature ($T{=}0.2$) that favors the top ranks while retaining some spread. Each task is a mutation that refines a single parent, or about $15\%$ of the time a crossover that samples a second parent from the same rank distribution and hands the worker both, so the two operations differ only in parent count. The policy then adds the scored child back, which our notation records as $\textsc{spawn}$, $\textsc{kill}$, and $\textsc{consolidate}$. Initialization is constant in the task, since every worker is born with the same blank context and uniform role before $\pi_{\text{meta}}$ assigns it a parent, so $\Init$ stays generic. A worker submits one candidate and then dies with no memory of the attempt, and the orchestrator scores that candidate in a separate container on a held-out split the worker never reads. Each task's data is split into three fixed parts, a training part the worker trains on, a search part that governs selection, and a validation part kept hidden from both the agents and the search until the run ends. This three-way hidden split is sandbox isolation and variance control, an expected protection, so $e$ stays generic. The budget $B$ is GPU-hours.

\vspace{-3mm}
\paragraph{AlphaEvolve ($A$, $S_{\text{cross}}$, $\pi_{\text{meta}}$).} AlphaEvolve is a coding agent for algorithmic discovery, such as faster matrix-multiplication algorithms, that evolves a program against a user-supplied objective \cite{novikov2025alphaevolve}. It shares AIRA$_2$'s evolutionary structure and sets the meta-control coordinate. AlphaEvolve sets the agent set $A$ through the backbone $\theta_i$ (various Gemini models) rather than through a count $|A|$. It pairs a cheaper Gemini~2.0~Flash that raises the candidate-generation rate for breadth with a stronger Gemini~2.0~Pro that supplies occasional higher-quality proposals for depth. The pairing gives two proposal distributions where AIRA$_2$ had one. An operation in $\Oset$ is a code edit, emitted as a search-and-replace diff block applied to a marked region of an existing program, or as a full rewrite when the block is short. The user marks the editable region, so the rest of the file stays a fixed skeleton. The calls exchange no direct messages, so the communication space $\Lambda=\emptyset$ as in AIRA$_2$, and the protocol $\sigma$ is generic. Every backbone may emit any edit, so the capability assignment $\alpha$ is uniform and generic. Coordination flows through the prompt the sampler builds from the store rather than a within-run workspace, so $S_{\text{btw}}=\emptyset$. The cross-run store $S_{\text{cross}}$ is a program database that keeps each program with its scores. It is organized around the MAP-Elites (i.e., a quality-diversity archive that keeps the best solution in each cell of a behavior grid) and island population models, so high scorers resurface as in-context exemplars for later prompts, and the store keeps diversity deliberately where AIRA$_2$ holds a single monolithic population. The meta-control policy $\pi_{\text{meta}}$ samples a parent and a set of high-scoring exemplars, biased toward both score and diversity through that organization, then routes the prompt to one of the two backbones. Each step builds the prompt, applies the diff, scores the child, and registers it back, which our notation records as $\textsc{spawn}$, $\textsc{kill}$, and $\textsc{consolidate}$ over the population. Initialization is the user-supplied starting program, evaluation code, and prompt template, problem-specific input rather than a coordinate the system sets, so $\Init$ is generic. The evaluator $e$ is programmatic and problem-specific, and it scores the artifact directly through a simulator, counter, or objective function. It runs as a cascade that drops weak candidates on a cheap stage before the expensive one, a smaller proxy gap than an LM-judge research-quality evaluator, so $e$ stays generic. The budget $B$ is compute-hours, where each solution can take on the order of 100 compute-hours spread asynchronously across a cluster, and thousands of LLM samples suffice where FunSearch needed millions \cite{novikov2025alphaevolve,romera2024mathematical}. The FunSearch-to-AlphaEvolve change is a scaling of the same evolutionary loop, from single short Python functions to entire multi-language files scored on several objectives, so the pair differs mainly in heterogeneous backbones and a diversity-keeping store over a shared evolutionary loop.

\vspace{-3mm}
\paragraph{Glia ($\alpha$).} Glia is a two agent supervisor-researcher system that writes a workload-adaptive request scheduling policy as Python code for a distributed GPU cluster serving large-language-model (LLM) inference \cite{hamadanian2025glia}. Its agent set is a fixed pair, $|A|=2$, which the authors call the Researcher and the Supervisor, both using o3 in the main evaluation. The operation universe $\Oset$ holds shell commands, file editing, and code execution. The two agents share one execution context, so the communication space $\Lambda$ is the bilateral edge set $\{(\mathrm{Res}\to \mathrm{Sup}),(\mathrm{Sup}\to \mathrm{Res})\}$, the protocol $\sigma$ is turn-taking dialogue, and $\pi_{\text{route}}$ alternates the two speakers. The capability assignment is asymmetric, and that asymmetry is the system's central design choice. The Researcher holds all of $\Oset$ and runs the simulator itself, navigating the codebase through standard Unix commands (\texttt{ls}, \texttt{grep}, \texttt{find}) in an agentic-search loop that proposes an idea, implements it, runs the experiment, and reads the result across turns. The Supervisor has $\alpha(\mathrm{Sup})=\emptyset$, no view of the codebase, and reads only the task description and the Researcher's outputs, from which it asks clarifying questions, recalls earlier findings, and redirects the Researcher toward overlooked directions, but it never edits code, runs the simulator, or proposes a ``concrete'' algorithm. The authors motivate the empty $\alpha(\mathrm{Sup})$ as keeping the Supervisor an independent reviewer rather than a second worker. The world state $S_{\text{world}}$ is a cloned copy of the serving codebase, and the within-run state $S_{\text{btw}}$ is the shared execution context. The cross-run state is empty, $S_{\text{cross}}=\emptyset$, so no learned strategy carries across runs, and even the multi-context variant, which runs $N$ independent single-context instances and returns the best, shares no history among them. The meta-control policy $\pi_{\text{meta}}$ is the identity, so no $\textsc{spawn}$ or $\textsc{grant}$ reshapes the configuration during a run. The evaluator $e$ is a simulator, which runs the edited policy over a fixed inference workload, a trace on four A10 GPUs serving Llama-3-8B, and returns a scalar, the mean request completion time. The Researcher does not mutate the simulator; it mutates the policy code the simulator then scores, so $e$ is generic. The same setup carries to batch scheduling and autoscaling, though the main case study designs a request router. The budget $B$ is the number of simulation runs, capped at 15 (relaxed to 100 for the multi-context variant), under an overall dollar cap of about \$30 per optimization.

\vspace{-3mm}
\paragraph{Automated Alignment Researcher (AAR) ($\Lambda$, $\Init$).} AAR searches for a training configuration that recovers a strong student's ground-truth-supervised performance from only a weak teacher's labels, the weak-to-strong supervision setting \cite{wen2026automatedw2s}. The agent set $A$ is nine parallel Claude~Opus~4.6 agents on a shared backbone $\theta_i$ and a uniform role $\rho_i$, each in its own sandbox running an autonomous ReAct loop with no prescribed workflow, which the authors found outperforms a fixed pipeline. The operation universe $\Oset$ holds stateful training and inference helpers, supplied baselines, and actions to submit for evaluation, share findings, and exchange codebases. The communication space $\Lambda$ is the broadcast access over a shared forum of findings held outside the sandboxes and synced into each one, so an agent browses the full set locally rather than querying a remote store by keyword, which the authors found recovers findings a keyword query would miss. The agents exchange no direct messages, so the message protocol $\sigma$ is the forum post alone. The capability assignment $\alpha$ is uniform, since $\alpha(a_i)=\Oset$ for every agent, so $\alpha$ stays generic. The within-run state $S_{\text{btw}}$ is the forum's contents, and nothing carries to a later run, so the cross-run state $S_{\text{cross}}=\emptyset$. The policy $\pi$ stays generic, since the meta-control policy $\pi_{\text{meta}}$ applies no population heuristic and each agent instead diagnoses its own failures from the training logs and reruns rather than abandoning the direction. The initialization $\Init$ seeds each agent with a distinct, deliberately ambiguous one-line research direction rather than a concrete idea, which sets $\Init$. Directed seeding ($\Init$) and temperature ($\eta$) both spread the agents across directions, and the tuple keeps them apart as distinct coordinates. The evaluator $e$ is a server-side scalar interface scored by performance gap recovered (PGR), the fraction of the strong-student gap a method recovers, with the agent submitting predictions to a remote interface whose labels never enter the sandbox. The budget $B$ is wall-clock time and dollars. AAR's directed-versus-undirected comparison isolates the $\Init$ contribution, holding $\theta_i$, $e$, and $B$ fixed (Section~\ref{sec:discussion}), and the scalar interface is weakened by exploits cataloged in Section~\ref{sec:objective}.

\vspace{-3mm}
\paragraph{AI Scientist-v2 ($\pi_{\text{meta}}$).} AI Scientist-v2 takes an open ML topic to a full manuscript with its supporting code, searching within a single run rather than across a population \cite{yamada2025aiscientistv2}. Its agent set $A$ is a single manager agent. The manager runs four sequential stages, from a working prototype through hyperparameter tuning, the research agenda, and ablations. Its private memory $m$ is the running agenda it carries across those stages (i.e., which stage is active and which node it handed forward) so it passes the membership test of Section~\ref{sec:mas} and keeps $|A|=1$. Within each stage a model call generates the plan and code for each new node. These calls hold no persistent private memory of their own, so they are modules in the mechanism rather than agents. The operations $\Oset$ are the Python code each call generates and runs in an interpreter, reading the metrics and logs back into the next call. The calls exchange no direct messages and coordinate only through the shared tree, so the communication space $\Lambda=\emptyset$ and the message protocol $\sigma$ is empty. Every call holds the same operations, so the capability assignment $\alpha$ is uniform and stays generic. The within-run state $S_{\text{btw}}$ is that tree, the only shared state, where a node holds an experiment script, its execution result, recorded metrics, and feedback. Nothing carries to a later run, so the cross-run state $S_{\text{cross}}=\emptyset$. The meta-control policy $\pi_{\text{meta}}$ is the $\textsc{fork}$-heavy rule that branches and prunes those nodes, the one coordinate the system sets, handing each stage's best node forward as the seed of the next. The signal that prunes is composite (e.g., code that errors marks a node buggy, a vision-language model gates the figures). This composite scorer is the selection rule inside $\pi_{\text{meta}}$, not an evaluator $e$ whose integrity is a separate choice, so $e$ stays generic. The initialization $\Init$ is the idea-generation phase that proposes a topic and checks its novelty against ``Semantic Scholar'' before the run begins. The system fixes this phase rather than varying the seeding as a design choice, and unlike AAR it runs no directed-versus-undirected comparison to single it out, so $\Init$ stays generic. The branching here is over experiments, not over ideas or drafts, which are generated linearly with a reflection pass. The budget $B$ is per-stage compute.

\vspace{-3mm}
\paragraph{MetaGPT ($A$, $\Lambda$, $\sigma$, $\pi_{\text{route}}$, $\Init$).} MetaGPT is a GPT-4 software-engineering team that turns a one-line requirement into a runnable codebase \cite{hong2024metagpt}. Its agent set $A$ is five fixed distinct members -- a product manager, an architect, a project manager, an engineer, and a quality-assurance engineer -- with the personas defined in $\Init$. The operations $\Oset$ are role-specific tools run in a ReAct loop, web search for the product manager, and code execution and debugging for the engineer, while the architect emits system-design diagrams and interface definitions as structured outputs rather than by invoking a tool. The communication space $\Lambda$ is not direct messaging. Each agent writes its structured documents to a shared store and reads the ones its role depends on, so the edges are the order those reads induce rather than a literal routing chain. The message protocol $\sigma$ is the fixed document format each role emits -- a requirements document, then a design with interfaces, then a task list, then code -- which the authors set against free-form dialogue to stop information from distorting across handoffs. The role-specific tools make the capability assignment differ across agents, yet they are bundled into each role and never varied on their own, so $\alpha$ is an artifact of the roles. That shared store is also the only within-run workspace. Its contents are the within-run state $S_{\text{btw}}$, which MetaGPT leaves generic, while the access order over it is the $\Lambda$ the system sets. Nothing carries to a later run, so $S_{\text{cross}}=\emptyset$. The routing policy $\pi_{\text{route}}$ is the fixed assembly-line order in which roles hand off, the coordinate that most separates MetaGPT from the evolutionary cases. The one piece of run-time adaptivity is a bounded retry loop in which the engineer reruns failing unit tests up to three times, which we read as a small $\pi_{\text{stop}}$ rule rather than a structural change, so $\pi_{\text{meta}}$ stays trivial. The personas are the initialization $\Init$, the per-role prompt fixing each agent's profile, goal, and constraints before the run, so $A$ sets how many distinct slots there are and $\Init$ sets what fills each. AAR sets the same axis with uniform agents instead. The evaluator $e$ is the unit-test pass rate on coding benchmarks, an inherited default rather than a coordinate it sets, so $e$ stays generic. The budget $B$ is tokens and dollars.

\vspace{-3mm}
\paragraph{EvoX ($\pi_{\text{meta}}$, $S_{\text{btw}}$).} EvoX is an evolutionary program-discovery system that rewrites its own search policy as the run proceeds \cite{evox2026}. The tasks are optimization problems such as circle packing and competitive-programming problems. The artifact is an evolved program scored by a task-specific programmatic evaluator that returns a scalar with auxiliary logs. The agent set is a pair of model calls, a solution generator and a strategy generator. An operation in $\Oset$ is a code edit applied under one of three variation operators, local refinement, structural variation, or free-form. The calls exchange no direct messages, so the communication space $\Lambda=\emptyset$ and the protocol $\sigma$ is vacuous. Either call may emit any edit, so the capability assignment $\alpha$ is uniform and generic. The object EvoX rewrites is the search strategy itself, the rule that builds the next prompt from the population by deciding which parent to mutate, which operator to apply, and which past solutions to include as inspiration. Rather than fixing $\pi$ at $t=0$, EvoX stores that strategy as a typed object in the within-run state $S_{\text{btw}}$, alongside a history of past strategies and their scores, and lets a separate language model emit a new version of it. Coordination stays within the run, so the cross-run state $S_{\text{cross}}=\emptyset$. The rewrite is demand-driven rather than periodic. When the best score stalls over a window, a programmatic meta-evaluator scores the spent strategy by the progress it produced, and the strategy model mutates a high-scoring past strategy into a replacement, while the solution population is preserved across the switch. In tuple terms EvoX sets $\pi_{\text{meta}}$, which emits $\textsc{rewrite-policy}(\pi')$, and the $S_{\text{btw}}$ that holds the policy object the rewrite acts on. Here $\pi$ becomes time-varying structural state rather than a fixed component, which is policy-as-state. Every run starts from the same uniform random search strategy, so $\Init$ is generic. The evaluator $e$ scores the artifact through a task-specific objective, an inherited default rather than a coordinate EvoX sets, so $e$ stays generic, and the meta-evaluator that scores strategies is internal to $\pi_{\text{meta}}$, not this $e$. The budget $B$ is a fixed cap of 100 evaluation iterations per task.

\vspace{-3mm}
\paragraph{ml-intern ($\pi_{\text{stop}}$).} ml-intern is a single-agent tool-calling loop for ML engineering, an open harness from Hugging Face \cite{mlintern2025}. With $|A|=1$, the harness is backbone-agnostic. The backbone $\theta$ is a launch-time choice, not a coordinate the system sets. The one agent runs its loop over operations $\Oset$ that read papers and datasets, edit files, run training jobs on local or sandboxed cloud compute, and evaluate the outcome. With a single agent the communication space $\Lambda$ is trivial, and the capability assignment $\alpha$ sends all of $\Oset$ to that one agent, so $\alpha$ is trivial as well. The within-run state $S_{\text{btw}}$ is the agent's single message history, kept by a context manager that auto-compacts at a token threshold, and nothing carries to a later run, so $S_{\text{cross}}=\emptyset$. The stopping policy $\pi_{\text{stop}}$ is the set coordinate, built from harness-level primitives. A doom-loop detector watches for repeated tool-call patterns and injects a corrective prompt when the run stalls, a hard cap of three hundred iterations is the terminal backstop, and the run also ends when the agent stops requesting tools. The meta-control policy $\pi_{\text{meta}}$ is trivial. Initialization $\Init$ is the user's one-line prompt, problem-specific input rather than a coordinate the system sets, so $\Init$ stays generic. ml-intern does not introduce a distinct external evaluator $e$. The agent interprets tool outputs inside its own loop, so $e$ is not a separately set coordinate in this system and remains generic. The budget $B$ is iterations, the same 300-step cap doubling as the terminal stop. The harness's helpers (e.g., the file, shell, and sandbox tools) are stateless and hold no private memory across calls, so by the membership test they are modules in $\Oset$ rather than members of $A$ (Section~\ref{sec:mas}).

\vspace{-3mm}
\paragraph{SimpleTES ($\pi_{\text{route}}$).} SimpleTES, for simple test-time evaluation-driven scaling, is a deliberate negative control on cross-run memory \cite{simpletes2026}. The task is an open scientific-discovery problem across six domains, including quantum-circuit compilation and GPU-kernel optimization, and the candidate solution is a program scored against a true objective the system cannot read directly. Its agent set is a single generator large language model (LLM), held to an open gpt-oss model to isolate the loop from generation-side scaling, so $|A|=1$. An operation produces one candidate solution from a constructed prompt. The capability assignment $\alpha$ and the communication space $\Lambda$ are trivial under a single agent. The within-run state $S_{\text{btw}}$ is the set of parallel refinement trajectories, and the system pins $S_{\text{cross}}=\emptyset$ at inference. It sets only $\pi_{\text{route}}$, a context constructor that selects which prior solutions to place in the next prompt. The selection is a graph version of predictor upper-confidence tree search (PUCT) that propagates value to a solution from the descendants it inspired and adds an exploration bonus for solutions rarely used as context, so the constructor favors prior solutions that seeded strong lineages. The replayed solutions are scoped to the current trajectory, narrower even than the run, since the parallel trajectories keep separate histories. Each trajectory starts from a baseline solution $y_0$, a problem-specific input rather than a coordinate the system sets, so $\Init$ is generic. The evaluator is a per-task programmatic surrogate $V$ for the true objective, an inherited default, so $e$ stays generic. The budget $B$ is evaluator queries, allocated across global width, refinement depth, and local sample size, which the authors scale jointly.

\vspace{-3mm}
\paragraph{Engram ($\pi_{\text{meta}}$, $S_{\text{cross}}$).} Engram inherits Glia's problem layer and fills the cross-run store $S_{\text{cross}}$ Glia leaves empty \cite{engram2026}. Engram's agent set $A$ is a sequential chain of single-active-agent explorations, each on a backbone $\theta_i$ that was o3, or gpt-5.2 in some runs. The operation universe $\Oset$, the communication space $\Lambda$, and the capability assignment $\alpha$ are all inherited from Glia's setting, with the same shell, file-editing, and code-execution tools. The state $S$ splits across two registers. The within-run state $S_{\text{btw}}$ is one agent's working context, reset at each handoff rather than shared. The cross-run state $S_{\text{cross}}$ is set, holding two objects on disk -- a per-experiment archive of code, scores, and logs, and a research digest of attempted approaches, insights, recommended next steps, and failed strategies, appended per agent -- which agents read on demand through tool calls rather than holding in context. Authors showed that ablating the digest costs more than ablating the archive, so the distilled reasoning helps successors more than the raw artifacts do. The set policy is $\pi_{\text{meta}}$. When an agent has spent its productive capacity, $\pi_{\text{meta}}$ emits $\textsc{consolidate}$, $\textsc{kill}$, and $\textsc{spawn}$ to distill its trajectory, discard its context window, and seed a fresh agent with an empty context. The seed $\Init$ stays generic, since each fresh agent receives the digest that $\textsc{spawn}$ hands it from $S_{\text{cross}}$ rather than a designed starting point of its own. The evaluator $e$ is task-dependent, inheriting Glia's simulator (Vidur) for the request-routing task, a cost verifier (egress plus VM cost) for its multi-cloud multicast task, and a prefix-hit-rate metric for its KV-cache task. The budget $B$ is 100 evaluation runs per task.

\section{Discussion}\label{sec:discussion}
In this study, we defined a shared vocabulary to describe the problem and architecture of a multi-agent autoresearch system, and illustrated that our vocabulary can describe and distinguish the designs of recently published autoresearch systems (Figure~\ref{fig:contrib-grid}). The same vocabulary extends beyond the case studies. For example, Shen et al.'s ``subagent'' architecture is a star $\Lambda$ in which parallel agents reach one another only through an orchestrator that $\textsc{join}$s their branches afterward, while their ``agent-team'' is a fixed chain in $\Lambda$ whose handoffs $\pi_{\text{route}}$ schedules before any run begins \cite{shen2026multiagentcollab}; Ueda et al.'s cohort size, interaction depth, and persona diversity resolve to $|A|$, dialogue length under $\pi_{\text{route}}$, and the spread of roles $\rho_i$ set by $\Init$ \cite{ueda2025ideation}.

Our vocabulary shows that existing work rarely isolates the bare agent count $|A|$, usually tying it to population size in $\pi_{\text{meta}}$, island count over $S_{\text{cross}}$, or a fixed value (e.g., $|A|=9$). The closest empirical probe is a standardized sweep across architectures and agent counts, which finds that \emph{architecture-task alignment} rather than agent count predicts gains \cite{kim2025scienceagentsystems}. Error amplification rises sharply in topologies that lack a centralizing verifier, which our vocabulary locates in $\Lambda$, $\pi_{\text{route}}$, and the verification integrity of $e$ rather than in $|A|$.

The run-time objective optimizes within a fixed $\Mset$, but the same vocabulary also frames the design question that precedes it, choosing which tuple to deploy for a task family, a budget, and an evaluator. For that question, the tuple makes ordinary ablations explicit, each holding all coordinates fixed except one. Vary $\Lambda$ to test communication topology. Vary $\alpha$ to test capability asymmetry. Strengthen $e$'s integrity while keeping $\pi_{\text{meta}}$ fixed to test whether apparent gains were evaluator artifacts. Compare $\Init$-based diversity against $\eta$-based diversity under the same budget. Some systems search this space automatically, for example over agent designs expressed as code or over workflow graphs~\cite{adas,aflow}, running the same proxy-optimization objective (Section~\ref{sec:objective}) over the space of tuples $\Mset$ rather than over the trajectories a fixed $\Mset$ explores.

The complaint that autoresearch systems lack \emph{taste} covers two distinct failures, and our vocabulary keeps them apart. \emph{Generative taste} is the rate at which the trajectory distribution $P_{\Mset}(\,\mathord{\cdot}\mid x)$ proposes novel trajectories before $e$ selects among them. It is a property of the generation path, set by the backbone priors $\theta_i$, the seeding $\Init$, the exploration term $\eta$, and what $S_{\text{cross}}$ has learned is worth trying. We leave generative taste qualitative without a novelty measure over $P_{\Mset}(\,\mathord{\cdot}\mid x)$ that would credit the generator the way $\Delta_\omega$ credits the evaluator. \emph{Evaluative taste} is the proxy-quality gap $\Delta_\omega$ between $e$ and $q$ (Section~\ref{sec:objective}). We believe a lack of novelty is fixed by changing $\Init$, $\eta$, or $\theta_i$, while a gamed metric is fixed by strengthening $e$'s integrity. For example, AAR's directed seeding raises generative taste by changing $R^{0}_{1:n}$, while its evaluator $e$ stays open to the exploits, a separate evaluative-taste failure~\cite{wen2026automatedw2s}. We should therefore attribute any reported gain to one or the other.

Generative taste sets how good the proposals are before selection, but $e$ chooses which proposal is returned. So, when $e$ is miscalibrated, harder search returns trajectories that score high on $e$ and low on $q$ regardless of how good the proposals were (i.e., the overfitting tax). We therefore suggest improving the evaluator's integrity before crediting a stronger generator. Crediting one means judging proposals by their novelty and quality under an evaluator decoupled from $e$, not by the final score. Integrity is one way to raise evaluative taste; $e$'s mechanism is another. When a task has ground-truth labels, a judge trained on them narrows $\Delta_\omega$ past what a prompted LLM judge reaches~\cite{su2026expertjudgment}. Since $q$ is uncomputable in general (Section~\ref{sec:problem}), it remains a target we design toward rather than a quantity we can measure.

Many coordinates remain that the field has not yet experimented with. Capability assignment is one. No system here changes $\alpha$ during a run, even though Glia exploits a non-trivial static $\alpha$. We take this as an engineering artifact, not evidence that dynamic capability control is useless. Dynamic $\alpha$ is held back by practical constraints, not by any argument against it, as prompt changes are cheaper than building the permission infrastructure that dynamic grants would require.

The control policy $\pi$ and the task distribution $\Dset$ contain three more under-exercised coordinates. The exploration term $\eta$ stays at its default, and even AAR injects diversity through $\Init$ rather than $\eta$. Run-time rewriting of $\pi$ itself is exercised by EvoX alone. A system that generates its own problems rather than solving a fixed set changes $\Dset$ (Section~\ref{sec:problem}), and no system here changes it. The pre-LLM POET co-evolved environments with the agents that solve them \cite{wang2019poet}, and in the LLM era CORAL co-evolves agents and their solutions but keeps the problem fixed \cite{qu2026coral}. A recent system begins to change $\Dset$, synthesizing its own tasks at a target difficulty, though it tunes only the solver, so the change stays partial \cite{autodata}. Beyond these, no action in $\Pi_{\text{meta}}$ changes the operation universe $\Oset$ or the backbone $\theta_i$ at all. A recursively self-improving successor rewrites its own $\Oset$ \cite{dgm} and updates its $\theta_i$ mid-run \cite{sia}. Both are already components of our tuple, so adding an operation-editing action over a mutable $\Oset_t$ and a weight-editing action over $\theta_{i,t}$ would let $\Pi_{\text{meta}}$ reach that case.

The clearest near-term test is on capability assignment. A senior advisor who reviews and redirects a junior researcher is useful partly because the advisor does not also write the code, run the experiments, or spend the budget -- the distance is what keeps the judgment independent. Hand the advisor the keyboard and the role collapses into a second worker. Glia builds exactly this distance into $\alpha$, giving its Supervisor $\alpha(\mathrm{Sup})=\emptyset$. The tuple turns the design into a one-coordinate ablation, holding $A$, $\Lambda$, $\pi_{\text{route}}$, $S_{\text{world}}$, and $e$ fixed while comparing the published static assignment against one that varies $\alpha$ during the run. A dynamic-$\alpha$ variant that improves score per dollar would identify dynamic capability assignment as useful, while one that spends more for no better score would credit the capability asymmetry rather than supervision alone. We expect the next round of systems to be distinguished by which of these coordinates they change, a difference the vocabulary makes explicit and a final score alone cannot.

\section*{Acknowledgments}
We thank Nicholas Dronen, Kevin Small, and Imry Kissos for their careful reading of the manuscript and for feedback that sharpened the argument.

\bibliographystyle{unsrt}
{\small
\bibliography{refs}
}

\appendix

\section{Related work}\label{sec:related}

We inherit our formulation from several literatures, generalize some of it, and depart from the rest.

\paragraph{Classical multi-agent systems.} The structural core of agents, communication, and shared state has a history in the multi-agent systems literature \cite{rao1995bdi,wooldridge2012introduction,shoham2008multiagent}. Those frameworks fix beliefs, desires, and intentions as the agent's internal state and treat communication as speech-acts over a typed message ontology. We keep the structural shape -- agents with private memory, a communication channel, a control policy -- but make two departures forced by the LLM-agent setting. First, the agent's internal state -- its backbone $\theta_i$, private memory $m_i$, and role $\rho_i$ -- is not a structured belief base, but a frozen language model plus an opaque context window, so propositional commitments and inference rules over them are not available. Second, the central object of analysis shifts from agent rationality to trajectory-distribution properties, meaning that what one can measure about an LLM-agent system is its run-time behavior, not its proof-theoretic commitments. The classical literature's abstractions remain useful for labeling components, and the present formalism uses them for that purpose, not for the soundness theorems they were designed to support.

\paragraph{Multi-agent reinforcement learning and Dec-POMDPs.} The trajectory-distribution formulation owes its shape to multi-agent reinforcement learning (MARL) and to decentralized partially observable Markov decision processes (Dec-POMDPs) \cite{bernstein2002complexity,littman1994markov}. From those frameworks, we borrow the run-as-trajectory view, the role of a stationary control policy, and the use of expected cumulative reward over a budget as the run-time objective. We deliberately do not borrow three things. First, MARL/Dec-POMDP formulations assume a transition kernel $P(s' \mid s, a)$ that is accessible for planning or learning; LLM-agent systems act on world state through operation calls whose effects are not modeled, so the tuple treats $S_{\text{world}}$ as black-box and the trajectory distribution $P_{\Mset}(\,\mathord{\cdot} \mid x)$ as the only object available. Second, MARL/Dec-POMDP fixes the agent population, action space, and observation space at $t=0$; we make structural change a primary action class via $\Pi_{\text{meta}}$, covering $\textsc{spawn}$, $\textsc{grant}$, and $\textsc{consolidate}$. Third, MARL/Dec-POMDP assumes a parameterized policy class amenable to gradient-based optimization; we permit this in principle but do not assume it. Most LLM-agent components are continuous in temperature and combinatorial in everything else. The closest contemporary relative is the literature on language-conditioned MARL (e.g., negotiation games and emergent communication), which shares the action-language interface but typically fixes the agent population and works in much smaller state spaces.

\paragraph{LLM-agent harnesses and surveys.} The recent wave of LLM-agent harnesses -- AutoGen, MetaGPT, CAMEL, ChatDev, AgentVerse, the Voyager skill-library line -- has produced a large taxonomy of design patterns such as planner/critic splits, supervisor-researcher dialogues, role-prompted teams, blackboard architectures, skill libraries, evolutionary populations \cite{wu2023autogen,hong2024metagpt,li2023camel,qian2023chatdev,chen2023agentverse,wang2023voyager}. A parallel line makes the harness itself the object of optimization. Some search over agent designs and workflows \cite{dgm,stop,adas,aflow}. Others evolve the cross-run store $S_{\text{cross}}$ our tuple names, accumulating it as an itemized ``playbook'' \cite{ace} or co-evolving the store with the routine that maintains it \cite{mce}. Survey papers organize these systems along axes such as communication topology, memory scheme, coordination protocol, and evolution mechanism \cite{gao2026survey,kim2025scienceagentsystems}. Gao et al.\ organize self-evolving agents around what evolves, when it evolves, and how adaptation is governed, including models, memory, tools, architecture, intra-test-time versus inter-test-time adaptation, and scalar-reward or textual-feedback mechanisms \cite{gao2026survey}. Kim et al.\ take a complementary empirical route, fitting quantitative scaling relationships across coordination pattern, model capability, system factors, and task factors while standardizing tools, prompts, and compute across hundreds of configurations \cite{kim2025scienceagentsystems}. Tie et al.\ survey this space under the name autoresearch and organize systems along a five-level human-to-AI autonomy spectrum, five workflow stages, and five evaluation dimensions \cite{tie2026autoresearchaiaipoweredresearch}. Their axes measure how much of the workflow the AI controls, executes, and validates, whereas we compare the agent structure and coordination of the systems themselves. These surveys catalog what a system looks like, but none gives the coordinates one holds fixed or varies to attribute an outcome to a single design choice. The tuple supplies those coordinates, separating cases that survey-level labels merge and recovering distinctions already present in the literature (Section~\ref{sec:discussion}).

\paragraph{Autoresearch systems.} We work through detailed examples later, which sit inside a larger autoresearch literature \cite{lu2026towards,yamada2025aiscientistv2,xia2026srscientist,qu2026coral,romera2024mathematical} (Section~\ref{sec:case-studies}). These systems do not define their designs in shared, comparable terms. They define them functionally, in prose about research threads, branches, memory, and a correctness audit, with a benchmark number attached. We use the tuple to label those design choices. As an example, one recent system runs many long-horizon research threads and combines promising branches, a meta-control search $\pi_{\text{meta}}$. It also consolidates context across experiments, a cross-run store $S_{\text{cross}}$ \cite{recursive2026}. Most autoresearch systems share both coordinates. What sets this system apart is the third coordinate. It strengthens $e$'s integrity against reward hacks inside the loop rather than fixing the evaluator in advance.

\paragraph{Reward hacking and Goodhart's law.} The alignment literature has long studied what happens when an agent optimizes a proxy in place of the quality it should match. Amodei et al.\ identify reward hacking as a safety problem, and Krakovna et al.\ catalog specification gaming across systems that met a stated objective while violating its intent \cite{amodei2016concrete,krakovna2020specification}. Manheim and Garrabrant decompose Goodhart's law into regressional, extremal, causal, and adversarial variants, separating benign proxy drift from optimization that seeks the gap \cite{manheim2018categorizing}. Skalse et al.\ define reward hacking formally and prove conditions under which a proxy is unhackable \cite{skalse2022defining}. Gao et al.\ fit scaling laws for how true reward diverges from proxy reward as optimization pressure grows \cite{gao2023rewardoveroptimization}, and Pang et al.\ show the same divergence in conditional text generation \cite{pang2023rewardgaming}. Wang and Huang treat the same problem as an equilibrium, proving that an agent optimizing a finite-dimensional evaluation signal must under-invest in the quality dimensions that signal omits \cite{wang2026rewardequilibrium}. The shared finding is that the gap widens with pressure and is a property of the proxy, not the optimizer's intent. We make two departures this literature does not. First, we identify the proxy with a single component, the evaluator $e$, so reward hacking becomes a property of $e$ rather than of the agents, and the distance from true quality is the proxy-quality gap (Section~\ref{sec:problem}). Second, we tie optimization pressure to how aggressively the meta-control policy $\pi_{\text{meta}}$ searches against $e$, so two systems with the same $e$ but different pressure occupy different points in the vocabulary. We then use both components to identify each documented hack as a specific integrity property of $e$ that failed (Section~\ref{sec:objective}).

\end{document}